\documentclass{article}

 \usepackage[main, final]{neurips_2026}

\usepackage{amsmath}
\usepackage{amssymb}
\usepackage{mathtools}
\usepackage{amsthm}
\usepackage{bm}
\usepackage{algorithm}
\usepackage{algorithmic}
\usepackage{graphicx}
\usepackage{subcaption}
\usepackage{multirow}
\usepackage{booktabs}
\usepackage{xcolor}
\usepackage{hyperref}
\usepackage{url}

\usepackage[capitalize,noabbrev]{cleveref}

\theoremstyle{plain}
\newtheorem{theorem}{Theorem}[section]
\newtheorem{proposition}[theorem]{Proposition}

\newtheorem{corollary}[theorem]{Corollary}
\theoremstyle{definition}
\newtheorem{definition}[theorem]{Definition}

\theoremstyle{remark}

\usepackage[textsize=tiny]{todonotes}

\title{Re-examining low-rank adaptation for private LLM fine-tuning}

\author{ Ali Dadsetan \& Frank Rudzicz \\ Dalhousie University \& Vector Institute \\ \texttt{\{ ali.dadsetan, frank\}@dal.ca} }

\begin{document}

\maketitle

\begin{abstract}
Privacy is a central concern when fine-tuning large language models (LLMs) on sensitive data, and differentially private stochastic gradient descent (DP-SGD) provides formal privacy guarantees. When performance at a fixed privacy level largely saturates on established benchmarks such as GLUE, we ask whether the same performance can be reached with less computation. Both theory and practice show that lower-rank models are better suited to DP training, a property especially relevant for LLMs, whose fine-tuning gradients exhibit a strong low-rank structure. Methods such as DP-LoRA exploit this by expressing each layer's weight update as a sum of only a few non-zero rank-one components. However, we argue that while having few non-zero components is important, the isotropic noise injected by DP-SGD inflates the singular values of the gradient matrix, disrupting their naturally fast decay. We investigate whether this noise-induced spectral distortion slows optimization. Experiments on language classification (GLUE with RoBERTa) and table-to-text generation (E2E and DART with Qwen and Llama models up to 4B parameters) show that partially restoring the original singular-value profile significantly accelerates DP-SGD convergence without compromising privacy guarantees. Our code is available at \url{https://github.com/alidadsetan/re-exam-private-low-rank}.
\end{abstract} 
\section{Introduction}
Many applications of machine learning to natural language processing raise privacy concerns because models trained on private data may leak sensitive information \citep{carlini2021extracting,carlini2022membership}.
These concerns extend beyond pre-training to fine-tuning on private data \citep{mireshghallah-etal-2022-empirical}.
Differential privacy (DP) \citep{dwork2014algorithmic} is a formal framework for quantifying and limiting the privacy loss experienced by individuals whose data are included in a dataset when an algorithm is applied to it.
DP-SGD \citep{abadi2016deep} provides privacy guarantees under the DP framework and has been successfully applied to fine-tuning LLMs \citep{yu2021differentially,li2021large,bu2022differentially,zhang2024disk}.

In particular, low-rank parameter-efficient methods have achieved remarkable success: LoRA-based private fine-tuning \citep{yu2021differentially}, bias-only tuning \citep{bu2022differentially}, and DiSK \citep{zhang2024disk} have each achieved state-of-the-art results for private LLM fine-tuning. The benefit of low model dimensionality for DP training is also supported theoretically \citep{nasr2021adversary,tran2025spectral}.

In DP-SGD, noise is deliberately added to the gradient vector before it is passed to the optimizer to ensure privacy. While this step is crucial for protecting individual data, it also slows down optimization (Figure~\ref{fig:clip_vs_dp} and Section~\ref{sec:slow-down}).
For transformer-based language models, gradient singular values typically decay rapidly, reflecting low matrix entropy and a strong low-rank structure \citep{li2022does,zhao2024galore}. DP noise slows this decay and increases matrix entropy \citep{li2022does}. To reduce computation on benchmarks where performance has largely saturated, we show that denoising methods from random matrix theory (RMT) can accelerate DP optimization by partially restoring the original singular-value structure. We validate this approach on classification tasks (GLUE \citep{wang2019glue} with RoBERTa \citep{liu2019roberta}) and generation tasks (E2E \citep{novikova2017e2e} and DART \citep{nan2021dart} with Qwen and Llama models).

A detailed discussion of changes in singular values under DP noise, including illustrative figures, is provided in Appendix~\ref{appendix:sv_distribution}.

\subsection{Related work}

Several works have addressed the computational overhead of differentially private training for language models. Approaches proposed by \citet{li2021large}, \citet{bu2023differentially}, and \citet{bu2022scalable} made the DP training of language models significantly faster. Parallel to these advances, \citet{yu2021differentially} and \citet{bu2022differentially}, among other works, made use of parameter-efficient DP fine-tuning of LLMs, achieving excellent performance with a small memory and compute budget.

Most closely related to our work, \citet{zhang2024doppler} and \citet{zhang2024disk} have proposed elegant denoising strategies that exploit the temporal correlation between gradients across consecutive training steps. DOPPLER \citep{zhang2024doppler} applies a low-pass filter over the gradient trajectory to attenuate the high-frequency noise introduced by DP-SGD, while DiSK-LoRA \citep{zhang2024disk} uses a Kalman-filter-inspired framework to combine past and current gradient estimates for more accurate updates.
However, these methods do not necessarily speed up optimization. Each DiSK step is twice as computationally expensive as a standard optimization step because it requires two backward passes, while DOPPLER is reported to slow early convergence.

Our work offers a complementary perspective. While the above methods denoise gradients across timesteps, we focus on a different question:
can we use the spectral structure of gradients, rather than their temporal structure, to design a denoising method that improves convergence speed?
To validate this insight, we apply spectral denoising independently at each training step to partially restore the original decay profile, and show that doing so yields measurable speedups. This per-step approach requires no gradient history, adds negligible overhead, and can in principle be combined with temporal methods.

\section{Background}

\subsection{Differential Privacy}
\label{sec:diff_privacy}
Differential privacy is a framework for quantifying the privacy risks of applying an algorithm to a sensitive training dataset. For a pair \((\epsilon,\delta)\), this formalism requires a learning algorithm \(\mathcal{M}\) to produce similar output distributions for two datasets differing in only one element. Intuitively, the output distribution should not change much depending on whether the algorithm sees a particular example. This intuition is formalized through approximate differential privacy.
\subsubsection{Approximate Differential Privacy}
\begin{definition}
    Two datasets are called neighboring if they differ only in the inclusion or exclusion of exactly one element.
\end{definition}
\begin{definition}
A randomized algorithm \(\mathcal{M}\) is said to satisfy \((\epsilon, \delta)\)-differential privacy if, for any two neighboring datasets \(D\) and \(D'\) and any event \(E\), the following holds:
\begin{equation}
\mathbb{P}(\mathcal{M}(D)\in E) \le \exp(\epsilon) \mathbb{P}(\mathcal{M}(D')\in E) + \delta
\end{equation}.

In practice, \(\delta\) is usually on the order of \(|D|^{-1}\) \citep{abadi2016deep}. In NLP applications, \(\epsilon\) usually takes values between \(0.5\) and \(8\) \citep{yu2021differentially,li2021large}.
\end{definition}
\subsubsection{DP-SGD}
\label{sec:dp-sgd}
DP-SGD is a popular method for training deep learning models with approximate differential privacy guarantees. It modifies the first-order SGD algorithm.

DP-SGD modifies the gradient before passing it to the optimizer through two steps: per-example gradient clipping and noise addition. These steps have two associated hyperparameters: the clipping threshold, \(C\), which controls the maximum per-example gradient norm, and the noise multiplier, \(\sigma\), which, when multiplied by \(C\), gives the standard deviation of the isotropic zero-mean Gaussian noise added to the sum of the clipped gradients \citep{abadi2016deep}.

In standard SGD, for a batch of data \(\{x_i\}_{i\in \mathcal{B}} \subset D\), the batch gradient is computed as:
\[
\bm{g}_{\mathcal{B}} = \frac{1}{B} \sum_{i\in \mathcal{B}} \bm{g}_i = \frac{1}{B} \sum_{i\in \mathcal{B}} \nabla f(\theta, x_i)
\]

In DP-SGD, each individual gradient is first clipped so that its norm does not exceed the threshold \(C\). The clipped gradients are then summed, and Gaussian noise with entries drawn from \(\mathcal{N}(0, \sigma_\text{DP}^2 C^2)\) is added. Finally, the result is averaged over the batch:

\begin{equation}
\label{eq:dp_sgd}
\begin{aligned}
\bar{\bm{g}}_{\mathcal{B}} &= \sum_{i\in \mathcal{B}} \mathrm{clip}(\bm{g}_i,C) \\
\tilde{\bm{g}}_{\mathcal{B}} &= \frac{1}{B}\left(\bar{\bm{g}}_{\mathcal{B}} + \bm{w}\right), \quad \bm{w}_{j} \sim \mathcal{N}(0, \sigma_\text{DP}^2 C^2)
\end{aligned}
\end{equation} 

This gradient is then passed to the optimizer of choice, e.g., SGD or Adam(W). The noise multiplier \(\sigma\) is selected based on the desired privacy guarantees \((\epsilon, \delta)\), the number of training steps, and the sampling rate (\(\frac{B}{|D|}\)). The required \(\sigma\) is computed using a privacy accountant. For this work, we use the privacy accountant of \citet{gopi2021numerical}, which provides tight privacy bounds.

% DP-SGD pseudocode using algorithm2e
\begin{algorithm}
\caption{DP-SGD (\textcolor{blue}{with Denoising})}
\label{alg:dp_sgd_denoise}
\begin{algorithmic}
\REQUIRE Dataset $D$, loss function $f(\theta, x)$, model parameters $\theta$, sampling rate $\rho$, clipping norm $C$, noise multiplier $\sigma$, optimizer $\mathcal{O}$, number of steps $T$, \textcolor{blue}{Denoising function $\mathrm{Denoise}(\cdot)$}
\FOR{$t \leftarrow 1$ to $T$}
   \STATE Sample a batch $\mathcal{B}$ from $D$ using Poisson sampling with rate $\rho$
   \FOR{each $i \in \mathcal{B}$}
      \STATE Compute per-example gradient $\bm{g}_i \leftarrow \nabla_\theta f(\theta, x_i)$
      \STATE Clip gradient: $\mathrm{clip}(\bm{g}_i, C) \leftarrow \bm{g}_i / \max(1, \|\bm{g}_i\|_2 / C)$
   \ENDFOR
   \STATE Aggregate clipped gradients: $\bar{\bm{g}} \leftarrow \sum_{i \in \mathcal{B}} \mathrm{clip}(\bm{g}_i, C)$
   \STATE Draw noise vector $\bm{w} \sim \mathcal{N}(0, \sigma_\text{DP}^2 C^2 \mathbf{I})$
   \STATE Compute noisy average: $\tilde{\bm{g}} \leftarrow (\bar{\bm{g}} + \bm{w}) / |\mathcal{B}|$
   \STATE \textcolor{blue}{\textbf{Denoise the gradient:} $\hat{\bm{g}} \leftarrow \mathrm{Denoise}(\tilde{\bm{g}})$}
   \STATE Update optimizer state: $\mathcal{O} \leftarrow \mathrm{UpdateState}(\mathcal{O}, \hat{\bm{g}})$
   \STATE Update parameters: $\theta \leftarrow \mathrm{UpdateParameters}(\theta, \mathcal{O})$
\ENDFOR
\end{algorithmic}
\end{algorithm}

\subsubsection{Post-processing invariance}
Post-processing a differentially private output without further access to private data preserves its privacy guarantees. Prior work uses this property to improve DP-SGD utility \citep{zhang2024doppler,pmlr-v80-balle18a}. To our knowledge, our method is the first to apply random matrix theory to post-process DP-SGD gradients.

\subsection{Singular value distribution of gradients}
\label{singular_value_decay}
The gradients of linear layers in neural networks exhibit a low-rank structure during training \citep{li2022does,zhao2024galore}. This structure is reflected in the rapid decay of the gradient matrix's singular values. This is a well-known phenomenon that has been observed in many settings \citep{li2022does,zhao2024galore}. In particular, \citet{li2022does} showed that the singular values of the gradient matrix decay exponentially and used this observation to explain why differential privacy works well in deep models with large parameter counts, contrary to theoretical expectations based on dimension alone.

However, prior work did not consider that the noise added by DP-SGD inflates the singular values of the gradient matrix, disrupting this fast decay and potentially slowing down the optimization process. In Figure~\ref{fig:roberta_svs}, we illustrate this effect by plotting the sorted singular values of a RoBERTa layer's gradient matrix before and after adding DP-SGD noise. Before noise is added, the singular values decay very rapidly, reflecting the strong low-rank structure of the gradient. After noise injection, the decay becomes markedly slower and the singular values do not approach zero until the very end of the spectrum.

\subsection{Random matrix theory background}
\label{sec:low_rank}
Low-rank matrix reconstruction is a rich subfield of signal processing \citep{donoho2018optimal,gavish2014optimal,shabalin2013reconstruction}. Assuming the rank of the signal matrix \(\bm{X} \in \mathbb{R}^{m \times n}\) is \(k\), we can use its SVD to write it as

\[
\bm{X} = \sum_{i=1}^k \lambda_i \bm{u}_i \bm{v}_i^T
\]

where the \(\lambda_i\) are singular values in non-increasing order, and the \(\bm{u}_i \in \mathbb{R}^m\) and \(\bm{v}_i \in \mathbb{R}^n\) form orthonormal sets of vectors.

We add a noise matrix with entries drawn from \(\mathcal{N}(0, \sigma^2)\) to obtain the noisy matrix \(\tilde{\bm{X}}\). The noise standard deviation \(\sigma\) differs from the noise multiplier \(\sigma_\text{DP}\) used in DP-SGD; they are related by \(\sigma = \frac{\sigma_\text{DP}C}{|\mathcal{B}|}\).

\[
\tilde{\bm{X}} = \bm{X} + \bm{\Delta}, \quad \bm\Delta_{ij} \overset{\text{i.i.d.}}{\sim} \mathcal{N}(0, \sigma^2)
\]

The goal is to estimate the original matrix \(\bm{X}\) from the noisy observation \(\tilde{\bm{X}}\). We write its SVD as

\[
\tilde{\bm{X}} = \sum_{i=1}^{\min(m, n)} \tilde{\lambda}_i \tilde{\bm{u}}_i \tilde{\bm{v}}_i^T
\]

The noisy matrix may (and usually does) have more than \(k\) nonzero components.

\subsubsection{Effect of noise on singular values and singular vectors: a phase transition}
\label{phase_transition}
With this notation, we have

\begin{equation}
\label{eq:sing_val}
\tilde{\lambda}_i \approx 
\begin{cases} 
F_{\sigma,n,m}(\lambda_i) & \text{if } \lambda_i > \sigma \sqrt[4]{mn} \\
\sigma (\sqrt{m} + \sqrt{n}) & \text{if } \lambda_i \leq \sigma \sqrt[4]{mn}
\end{cases}
\end{equation}
where
\begin{equation*}
F_{\sigma,n,m}(\lambda) = \sqrt{\left(\lambda + \frac{\sigma^2 n}{\lambda}\right)\left(\lambda + \frac{\sigma^2 m}{\lambda}\right)}
\end{equation*}
This noise-induced increase in singular values is a standard phenomenon in random matrix theory. Note that by Equation~\ref{eq:sing_val}, even the first few singular values---which under the low-rank assumption carry the signal---are inflated to at least \(F_{\sigma,n,m}(\lambda_i)\). Since \(F_{\sigma,n,m}(\lambda) > \lambda\) for all \(\lambda > 0\), every detectable singular value is strictly blown up. This minimum inflation is the source of the singular-value blow-up discussed in the introduction.

\paragraph{Effect of noise on singular vectors.}
Noise affects both the singular values and the singular vectors. Assuming the nonzero singular values of \(\bm{X}\) are distinct and \(\lambda_i > \sigma\sqrt[4]{mn}\), following Lemma 3 of \citet{gavish2014optimal} or Proposition 9 of \citet{shabalin2013reconstruction}, we can write
\begin{equation}
\label{eq:sim_u}
\left| \langle \bm{u}_i, \, \tilde{\bm{u}}_{j} \rangle \right|^2 
\approx 
\begin{cases}
\displaystyle \frac{\lambda_i^4 - mn\sigma^4}{\lambda_i^4 + m\lambda_i^2\sigma^2} & i = j \\
0 & i \neq j
\end{cases},
\end{equation}

and

\begin{equation}
\label{eq:sim_v}
\left| \langle \bm{v}_i, \, \tilde{\bm{v}}_{j} \rangle \right|^2 
\approx 
\begin{cases}
\displaystyle \frac{\lambda_i^4 - mn\sigma^4}{\lambda_i^4 + n\lambda_i^2\sigma^2} & i = j \\
0 & i \neq j
\end{cases}.
\end{equation}

However, if \(\lambda_i \leq \sigma\sqrt[4]{mn}\), then

\begin{equation}
\left| \langle \bm{u}_i, \, \tilde{\bm{u}}_{j} \rangle \right|^2 
\approx \left| \langle \bm{v}_i, \, \tilde{\bm{v}}_{j} \rangle \right|^2 
\approx 0
\end{equation}

In other words, in the asymptotic limit, singular vectors corresponding to singular values below the phase-transition threshold carry no useful information about the original signal---they are indistinguishable from random directions.

\paragraph{Applicability to modern LLMs.}
Although the above results are derived in the large-matrix asymptotic regime, they are well suited to the DP fine-tuning setting. While LoRA restricts the update rank to a small value, each of its two low-rank factors ($A$ and $B$) has one dimension equal to the hidden width of the network, which in modern LLMs is large (often in the thousands). Consequently, the gradient matrices associated with each factor are inherently high-dimensional and the asymptotic predictions are accurate in practice. More broadly, analyzing neural networks through the lens of infinite-width or large-dimensional limits has proven remarkably insightful across a range of settings \citep{jacot2018neural,lee2019wide,mei2018mean,blake2024u}.

\subsubsection{Matrix Denoising}
\label{sec:matrix_denoising}
Given the low-rank signal model described above, matrix denoising methods can partially recover the original signal matrix \(\bm{X}\) from its noisy observation \(\tilde{\bm{X}}\) by shrinking the singular values of the noisy matrix \citep{shabalin2013reconstruction,donoho2018optimal}. In particular, the so-called optimal estimator of \citet{shabalin2013reconstruction} produces a low-rank output using only the SVD of the noisy matrix and knowledge of the noise variance \(\sigma^2\). The computation adds negligible overhead to training: with an efficient, batched implementation the denoising step accounts for less than \(1\%\) of total training time in our experiments. Full derivations, formulas, and a computational-overhead comparison are provided in Appendix~\ref{appendix:matrix_denoising}.

\section{Noise causes optimization slowdown}
\label{sec:slow-down}
\begin{figure}[h]
   \centering
   \includegraphics[width=0.6\linewidth]{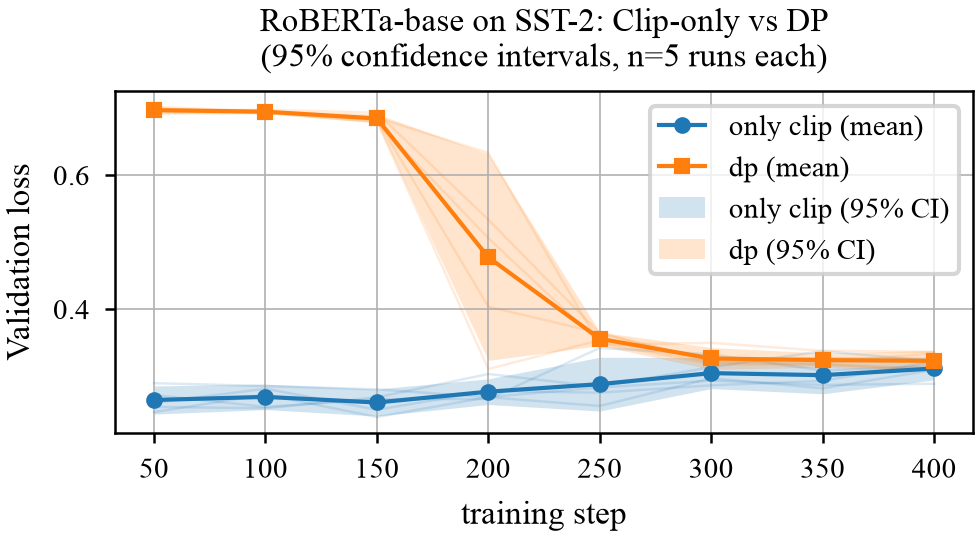}
   \caption{Comparison of validation loss between DP-SGD and training with per-sample clipping only (no noise added), using RoBERTa-base on SST-2 with clipping threshold \(C=10\). Settings follow Section~\ref{sec:experiments}, except that clipping-only training omits Gaussian noise. The slowdown in convergence is more pronounced for DP-SGD than for clipping alone. Lines show the mean of five runs; shaded regions show 95\% confidence intervals computed from standard errors.}
   \label{fig:clip_vs_dp}
\end{figure}

In this section, we study the convergence slowdown caused by adding Gaussian noise to clipped gradients. Although prior work often attributes the loss gap between DP-SGD and non-private training to clipping \citep{bu2023convergence}, convergence speed and final performance are distinct phenomena. For \(C=10\), Figure~\ref{fig:clip_vs_dp} shows that the final validation loss is similar for DP-SGD with a zero noise multiplier (\(\sigma=0\)) and for DP-SGD using the noise level required for privacy; however, convergence toward that final loss is significantly slower when noise is added. This comparison highlights the effect of noise on convergence speed while holding the clipping configuration fixed. Additional comparisons of clipping-only training and DP-SGD across datasets appear in Appendix~\ref{appendix:clip_vs_dp}.

\subsection{Sensitivity to clipping configuration}
\label{sec:clipping_sensitivity}
\begin{figure}[htbp]
   \centering
   \includegraphics[width=\linewidth]{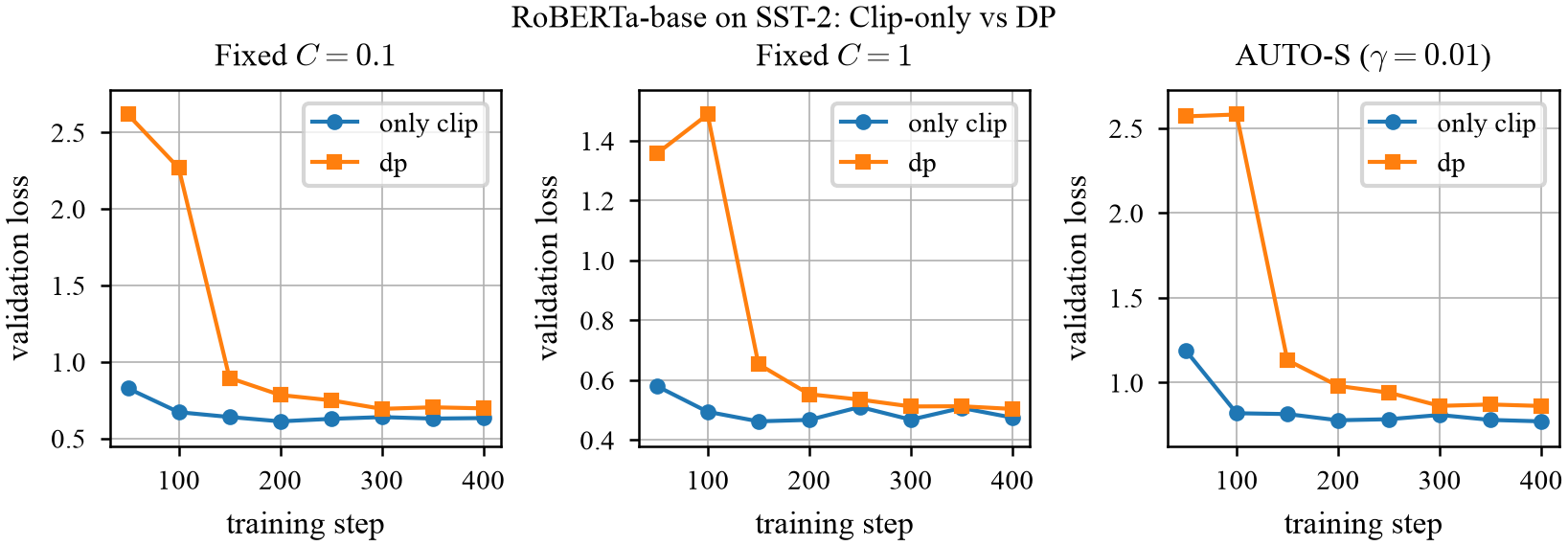}
   \caption{Effect of the clipping configuration on validation loss for RoBERTa-base on SST-2. Each panel compares clipping-only training with a matched DP-SGD run for fixed \(C=0.1\), fixed \(C=1\), or AUTO-S (\(\gamma=0.01\)). Lines connect the eight reported checkpoints. Each panel uses its own vertical scale. The \(C=10\) comparison appears in Figure~\ref{fig:clip_vs_dp}.}
   \label{fig:clipping_threshold_ablation}
\end{figure}

To test whether the observed slowdown depends on the clipping configuration, we compare clipping-only training with matched DP-SGD runs on SST-2 using RoBERTa-base under fixed thresholds \(C\in\{0.1,1,10\}\) and automatic clipping (AUTO-S with \(\gamma=0.01\)) \citep{bu2023automatic}. Figure~\ref{fig:clipping_threshold_ablation} shows the additional clipping settings; the \(C=10\) comparison appears in Figure~\ref{fig:clip_vs_dp}. Across all four configurations, DP-SGD has higher validation loss at every reported checkpoint from step 50 to step 400, with the largest gaps at step 50 or 100. The gaps narrow later in training, supporting a noise-induced convergence slowdown across the tested clipping settings. These results do not establish that noise is the primary cause of final-performance gaps, which may also reflect clipping.

\section{Methodology}
\label{sec:methodology}
In this section, we introduce our post-processing method, which leverages equations~\ref{eq:optimal_denoising} and~\ref{eq:optimal_eta} to denoise the gradients produced by DP-SGD before they are passed to the optimizer. 

\subsection{Denoising Function}
\label{sec:denoising_function}
We apply the matrix denoising method from Appendix~\ref{appendix:matrix_denoising} independently to each linear layer of the network. For a linear layer $\bm{W}_i$, the restriction of the noisy gradient to that layer, $\tilde{\bm{g}}\!\mid_{\bm{W}_i}$, is treated as a noisy low-rank matrix and denoised via the optimal estimator. Non-linear layers are left unchanged. The full-gradient denoising is then the concatenation of per-layer outputs:
\[
\mathrm{Denoise}(\tilde{\bm{g}}) = \left(\mathrm{Denoise}(\tilde{\bm{g}}\!\mid_{W_1}),  \ldots, \mathrm{Denoise}(\tilde{\bm{g}}\!\mid_{W_L})\right).
\]

Two practical modifications are required to make this layer-wise strategy effective:
\begin{enumerate}
   \item \textbf{Thresholding.} We only denoise a layer if its largest singular value exceeds $\kappa\,\sigma(\sqrt{n} + \sqrt{m})$, where $n, m$ are the layer dimensions and $\kappa = 1.02$. This avoids applying the estimator in the regime where the signal is indistinguishable from noise and the optimal estimate is the zero matrix. 
   \item \textbf{Norm correction.} After denoising, we rescale the output so that $\|\mathrm{Denoise}(\tilde{\bm{g}}\!\mid_{\bm{W}_i})\| = \|\tilde{\bm{g}}\!\mid_{\bm{W}_i}\|$. This ensures that per-layer cosine-similarity improvements (justified in the asymptotic setting in Appendix~\ref{appendix:why_optimal_denoising}) translate into an improvement for the full gradient vector (Theorem~\ref{theorem:norm_correction} in Appendix~\ref{appendix:denoising_details}).
\end{enumerate}
Detailed justifications for both choices, including ablation studies and the formal theorem, are provided in Appendix~\ref{appendix:denoising_details}.

\begin{figure}[h]
   \centering
   \includegraphics[width=.6\linewidth]{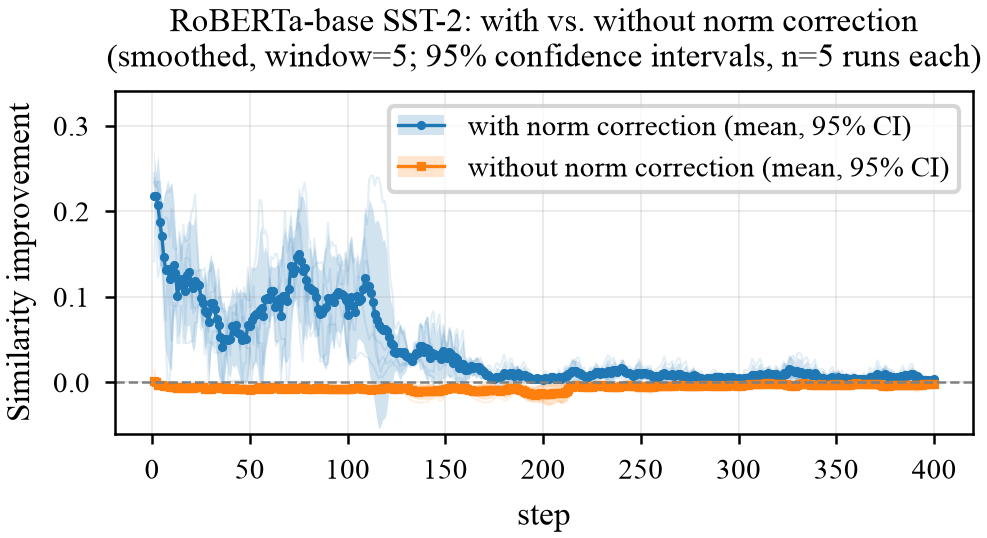}
   \caption{Comparison of whole-gradient cosine-similarity improvement when using norm correction versus not using it, for RoBERTa-base on SST-2. With norm correction (our method), the mean improvement is positive throughout training. Curves are smoothed with a five-step moving average; lines show the mean of five runs, and shaded regions show 95\% confidence intervals computed from standard errors.}
   \label{fig:norm_abal_sst_base}
\end{figure}

To evaluate how denoising affects alignment with the clipped gradient, we define the improvement at step \(t\) as
\[
\mathrm{Improvement}(t) = \mathrm{cos}(\mathrm{Denoise}(\tilde{\bm{g}}_t), \bar{\bm{g}}_t) - \mathrm{cos}(\tilde{\bm{g}}_t, \bar{\bm{g}}_t).
\]
Here, \(\bar{\bm{g}}_t\) and \(\tilde{\bm{g}}_t\) are the clipped and noisy gradients, respectively, and \(\mathrm{cos}(\bm{a},\bm{b}) = \frac{\bm{a}^{\top}\bm{b}}{\|\bm{a}\|_2\|\bm{b}\|_2}\) denotes cosine similarity. This metric is used only for evaluation; the algorithm denoises only already-privatized gradients (Algorithm~\ref{alg:dp_sgd_denoise}), preserving the privacy guarantees of DP-SGD.
Figure~\ref{fig:norm_abal_sst_base} confirms that the resulting denoising function achieves a positive mean cosine-similarity improvement between the denoised and clipped gradients throughout training.

\paragraph{Why is higher cosine similarity desirable?}
Consider two update directions \(\Delta\theta_1, \Delta\theta_2\) with \(\|\Delta\theta_1\| = \|\Delta\theta_2\| = r\), where \(\Delta\theta_1\) has higher cosine similarity with the gradient \(g = \nabla L(\theta)\). A gradient-descent step \(\theta \mapsto \theta - \alpha\,\Delta\theta\) gives the one-step loss difference
\[
L(\theta - \alpha\,\Delta\theta_1) - L(\theta - \alpha\,\Delta\theta_2)
\;=\; -\alpha\,g^{\!\top}(\Delta\theta_1 - \Delta\theta_2) + \mathcal{O}(\alpha^2).
\]
Since \(\cos(\Delta\theta_1, g) > \cos(\Delta\theta_2, g)\) and both directions share the same norm, we have \(g^{\!\top}\Delta\theta_1 > g^{\!\top}\Delta\theta_2\), so the leading term is negative. For sufficiently small \(\alpha\) the \(\mathcal{O}(\alpha^2)\) remainder is dominated, giving \(L(\theta - \alpha\,\Delta\theta_1) < L(\theta - \alpha\,\Delta\theta_2)\): the better-aligned direction yields a larger loss decrease. A formal statement and proof for the Adam optimizer (under a quadratic model) are given in Appendix~\ref{appendix:adam_lower_bound}. A discerning reader may have noticed that our denoising targets the clipped gradient, whereas the argument above concerns the true gradient. Appendix~\ref{appendix:unclipped_alignment} examines this distinction empirically, reporting positive alignment improvements with the unclipped gradient for RoBERTa-base on SST-2 and MNLI.

\section{Experiments}
\label{sec:experiments}

In this section, we present the evaluation of our proposed method. To assess our goal of improving the convergence speed of DP-SGD, we compared the performance of DP-SGD with and without our SVD denoising method across four classification datasets from the GLUE benchmark \citep{wang2019glue} and two model scales of the RoBERTa family \citep{liu2019roberta}. The baseline method is standard DP-LoRA. We provide the full experimental details and hyperparameter settings for all experiments in Appendix~\ref{appendix:experimental_details}.

We report step speedup as \((T-T_{\text{ours}})/T\), where \(T_{\text{ours}}\) is the linearly interpolated step matching the baseline accuracy at step \(T\). We denote the corresponding privacy-budget saving by \(\Delta\epsilon\).

Table~\ref{tab:glue_results} shows the results. The method with the corrected singular-value structure performed better than the baseline on average across steps, while converging significantly faster. Consequently, our approach successfully reaches the same accuracy while consuming a much smaller privacy budget (\(\epsilon\)). At step 200, the step speedups range from \textbf{15.6\% to 75.0\%}, with corresponding \(\epsilon\) savings of up to 1.64. At step 400, the step speedups range from \textbf{15.6\% to 47.9\%}. In the single case (QNLI Large at step 400) where the mean accuracy of our method is nominally lower than the baseline, the difference lies within the 95\% confidence interval of the baseline and is not statistically significant (Welch's $t$-test, $p=0.26$), confirming that our method does not degrade performance even when it matches the baseline's convergence.

\begin{table}[ht]
\centering
\setlength{\tabcolsep}{4pt}
\renewcommand{\arraystretch}{1.1}
\resizebox{\columnwidth}{!}{%
\begin{tabular}{ll cccc cccc}
\toprule
\multirow{2}{*}{Task} & \multirow{2}{*}{Model} & \multicolumn{4}{c}{Results at Step 200} & \multicolumn{4}{c}{Results at Step 400} \\
\cmidrule(lr){3-6} \cmidrule(lr){7-10}
 & & DP-LoRA Acc. & Ours Acc. & Speedup & $\Delta\epsilon$ & DP-LoRA Acc. & Ours Acc. & Speedup & $\Delta\epsilon$ \\
\midrule
\multirow{2}{*}{SST-2} 
& Base & $87.9 \pm 1.2$ & \textbf{90.8} $\pm$ 0.3 & 27.3\% & 0.57 & $91.8 \pm 0.3$ & \textbf{92.1} $\pm$ 0.3 & 17.5\% & 0.55 \\
& Large & $65.5 \pm 24.8$ & \textbf{93.1} $\pm$ 0.2 & 62.5\% & 1.47 & $94.2 \pm 0.5$ & \textbf{94.2} $\pm$ 0.2 & 15.6\% & 0.49 \\
\midrule
\multirow{2}{*}{MNLI}
& Base & $41.1 \pm 6.3$ & \textbf{71.2} $\pm$ 3.5 & 46.4\% & 0.55 & $77.0 \pm 0.2$ & \textbf{79.9} $\pm$ 0.5 & 34.5\% & 0.56 \\
& Large & $62.4 \pm 10.7$ & \textbf{78.8} $\pm$ 8.3 & 15.6\% & 0.17 & $85.3 \pm 0.5$ & \textbf{85.9} $\pm$ 0.1 & 19.3\% & 0.30 \\
\midrule
\multirow{2}{*}{QNLI}
& Base & $50.3 \pm 2.3$ & \textbf{78.2} $\pm$ 1.3 & 75.0\% & 1.64 & $80.7 \pm 0.9$ & \textbf{84.4} $\pm$ 0.3 & 40.5\% & 1.17 \\
& Large & $83.2 \pm 0.9$ & \textbf{85.1} $\pm$ 0.3 & 21.7\% & 0.39 & \textbf{88.8} $\pm$ 0.8 & $88.4 \pm 0.3$ & N/A & N/A \\
\midrule
\multirow{2}{*}{QQP}
& Base & $70.2 \pm 7.2$ & \textbf{81.7} $\pm$ 0.2 & 43.7\% & 0.53 & $81.7 \pm 0.1$ & \textbf{83.2} $\pm$ 0.1 & 47.9\% & 0.83 \\
& Large & $75.8 \pm 1.2$ & \textbf{82.8} $\pm$ 0.4 & 35.0\% & 0.41 & $84.0 \pm 0.2$ & \textbf{84.7} $\pm$ 0.4 & 27.9\% & 0.45 \\
\bottomrule
\end{tabular}%
}
\vspace{0.1cm}
\caption{Comparison of our SVD denoising method with standard DP-LoRA on GLUE classification tasks. We report mean accuracy across runs $\pm$ the half-width of the 95\% confidence interval, computed from the standard error over runs (Appendix~\ref{appendix:experimental_details}). Step speedup and $\Delta\epsilon$ (privacy budget saved) represent the percentage of iterations and the absolute privacy budget saved by reaching the baseline's accuracy at a given step milestone. Speedup is calculated via linear interpolation.}
\label{tab:glue_results}
\end{table}

\paragraph{Robustness to clipping configuration.}
We compare our method with DP-LoRA on SST-2 using RoBERTa-base under three fixed clipping thresholds and AUTO-S. All runs last 400 steps; Table~\ref{tab:clipping_configuration} reports the first three validation checkpoints. Our method matches DP-LoRA at step 50 and achieves higher accuracy at steps 100 and 150 in all four settings, supporting faster early optimization across the tested clipping configurations.

\begin{table}[h]
\centering
\small
\setlength{\tabcolsep}{5pt}
\renewcommand{\arraystretch}{1.1}
\begin{tabular}{l rr rr rr}
\toprule
\multirow{2}{*}{Clipping method} & \multicolumn{2}{c}{Step 50} & \multicolumn{2}{c}{Step 100} & \multicolumn{2}{c}{Step 150} \\
\cmidrule(lr){2-3} \cmidrule(lr){4-5} \cmidrule(lr){6-7}
 & DP-LoRA & Ours & DP-LoRA & Ours & DP-LoRA & Ours \\
\midrule
Fixed \(C=0.1\) & 50.92 & 50.92 & 50.92 & \textbf{89.45} & 89.89 & \textbf{90.30} \\
Fixed \(C=1\) & 50.92 & 50.92 & 50.92 & \textbf{89.54} & 89.54 & \textbf{90.73} \\
Fixed \(C=10\) & 50.92 & 50.92 & 50.92 & \textbf{80.96} & 50.92 & \textbf{90.14} \\
AUTO-S (\(\gamma=0.01\)) & 50.92 & 50.92 & 50.92 & \textbf{89.36} & 89.29 & \textbf{90.64} \\
\bottomrule
\end{tabular}
\vspace{3pt}
\caption{Validation accuracy (\%) on SST-2 with RoBERTa-base across clipping configurations. Bold indicates the higher accuracy within each pair.}
\label{tab:clipping_configuration}
\end{table}

\paragraph{Comparison with other denoising methods.}
DOPPLER \citep{zhang2024doppler} and DiSK \citep{zhang2024disk} primarily target privacy-noise reduction and improved utility; we include comparisons with these methods for completeness in our evaluation of early convergence. We evaluate both methods combined with DP-LoRA on SST-2 and MNLI using RoBERTa-base and RoBERTa-large. Because their per-step costs differ, we compare at equal wall-clock budgets corresponding to 200 and 400 DP-LoRA steps (Table~\ref{tab:denoising_baselines}). Our method achieves the highest reported mean accuracy in all eight comparisons while completing nearly as many updates as DP-LoRA; DiSK completes roughly half as many within the same budgets. DP-LoRA remains the most direct baseline for isolating our denoiser's contribution. Computational overhead is analyzed further in Appendix~\ref{appendix:matrix_denoising}, Table~\ref{tab:computational_overhead}.

\begin{table}[h]
\centering
\small
\setlength{\tabcolsep}{3pt}
\renewcommand{\arraystretch}{1.1}
\resizebox{\columnwidth}{!}{%
\begin{tabular}{llr rr rr rr rr}
\toprule
\multirow{2}{*}{Task} & \multirow{2}{*}{Model} & \multirow{2}{*}{\shortstack{Budget\\(min)}} & \multicolumn{2}{c}{DP-LoRA} & \multicolumn{2}{c}{DOPPLER} & \multicolumn{2}{c}{DiSK} & \multicolumn{2}{c}{Ours} \\
\cmidrule(lr){4-5} \cmidrule(lr){6-7} \cmidrule(lr){8-9} \cmidrule(lr){10-11}
 & & & Steps & Acc. (\%) & Steps & Acc. (\%) & Steps & Acc. (\%) & Steps & Acc. (\%) \\
\midrule
\multicolumn{11}{l}{\textit{Time budget corresponding to 200 DP-LoRA steps}} \\
\addlinespace[2pt]
\multirow{2}{*}{MNLI}
 & Base & 26.79 & 200 & $41.74\pm6.41$ & 200 & $37.48\pm6.02$ & 97 & $34.95\pm2.52$ & 197 & $\mathbf{70.71}\pm2.75$ \\
 & Large & 77.34 & 200 & $63.38\pm7.28$ & 198 & $52.73\pm8.43$ & 96 & $36.24\pm2.07$ & 197 & $\mathbf{78.81}\pm2.21$ \\
\addlinespace[2pt]
\multirow{2}{*}{SST-2}
 & Base & 26.74 & 200 & $85.28\pm6.83$ & 199 & $82.85\pm8.16$ & 97 & $50.92\pm0.00$ & 197 & $\mathbf{90.95}\pm0.43$ \\
 & Large & 76.98 & 200 & $64.15\pm22.65$ & 198 & $72.83\pm21.87$ & 96 & $50.92\pm0.00$ & 196 & $\mathbf{93.18}\pm0.30$ \\
\midrule
\multicolumn{11}{l}{\textit{Time budget corresponding to 400 DP-LoRA steps}} \\
\addlinespace[2pt]
\multirow{2}{*}{MNLI}
 & Base & 53.59 & 400 & $77.04\pm0.57$ & 399 & $75.19\pm1.39$ & 195 & $54.21\pm4.80$ & 394 & $\mathbf{79.77}\pm0.46$ \\
 & Large & 154.74 & 400 & $85.38\pm0.29$ & 396 & $84.70\pm0.73$ & 192 & $75.44\pm3.59$ & 394 & $\mathbf{85.69}\pm0.28$ \\
\addlinespace[2pt]
\multirow{2}{*}{SST-2}
 & Base & 53.45 & 400 & $91.65\pm0.27$ & 399 & $91.92\pm0.24$ & 195 & $88.12\pm2.50$ & 394 & $\mathbf{92.23}\pm0.22$ \\
 & Large & 153.96 & 400 & $93.97\pm0.08$ & 397 & $94.01\pm0.27$ & 191 & $86.02\pm4.04$ & 394 & $\mathbf{94.07}\pm0.30$ \\
\bottomrule
\end{tabular}%
}
\vspace{3pt}  
\caption{Comparison under equal wall-clock budgets on MNLI and SST-2 with RoBERTa-base and RoBERTa-large. DOPPLER, DiSK, and our method are each combined with DP-LoRA. We report steps completed and mean accuracy with 95\% confidence intervals computed from standard errors. Accuracy is evaluated every 50 steps and linearly interpolated between checkpoints. Bold marks the highest mean accuracy in each row.}
\label{tab:denoising_baselines}
\end{table}

\paragraph{Language generation.} We also report language-generation experiments on the E2E benchmark \citep{novikova2017e2e} and DART \citep{nan2021dart} using Qwen and Llama-3.2 models \citep{qwen3technicalreport,grattafiori2024llama}. After 50 training steps (Table~\ref{tab:e2e_dart_results_new_50}), our denoising method performs at least as good as the baseline in \textbf{49 of 50} model/dataset/metric combinations, demonstrating substantial sample-efficiency gains under tight iteration budgets. These experiments use models of up to 4B parameters, indicating that the approach scales well to larger models. Detailed experimental setups for these generation tasks are also provided in Appendix~\ref{appendix:experimental_details}.

\begin{table}[ht]
\centering
\setlength{\tabcolsep}{4pt}
\renewcommand{\arraystretch}{1.1}
\resizebox{\columnwidth}{!}{%
\begin{tabular}{ll cc cc cc cc cc}
\toprule
\multirow{2}{*}{Task} & \multirow{2}{*}{Model} & \multicolumn{2}{c}{BLEU} & \multicolumn{2}{c}{ROUGE-L} & \multicolumn{2}{c}{METEOR} & \multicolumn{2}{c}{NIST} & \multicolumn{2}{c}{CIDEr} \\
\cmidrule(lr){3-4} \cmidrule(lr){5-6} \cmidrule(lr){7-8} \cmidrule(lr){9-10} \cmidrule(lr){11-12}
 & & DP-LoRA & Ours & DP-LoRA & Ours & DP-LoRA & Ours & DP-LoRA & Ours & DP-LoRA & Ours \\
\midrule
\multirow{5}{*}{E2E} & Qwen 0.6B & 23.17 & \textbf{37.35} & 46.78 & \textbf{53.52} & 0.509 & \textbf{0.628} & 2.34 & \textbf{4.60} & 0.67 & \textbf{1.11} \\
 & Qwen 1.7B & 36.01 & \textbf{36.95} & 53.57 & \textbf{54.42} & 0.609 & \textbf{0.624} & 4.64 & \textbf{4.71} & 1.08 & \textbf{1.14} \\
 & Qwen 4B & 40.00 & \textbf{40.88} & 55.19 & \textbf{55.74} & 0.659 & \textbf{0.663} & 4.83 & \textbf{5.03} & 1.27 & \textbf{1.36} \\
 & Llama 1B & 22.15 & \textbf{36.33} & 44.24 & \textbf{52.86} & 0.476 & \textbf{0.626} & 2.59 & \textbf{4.62} & 0.47 & \textbf{1.13} \\
 & Llama 3B & 10.96 & \textbf{26.33} & 25.22 & \textbf{44.36} & 0.268 & \textbf{0.516} & 0.93 & \textbf{3.48} & 0.22 & \textbf{0.64} \\
\midrule
\multirow{5}{*}{DART} & Qwen 0.6B & 14.66 & \textbf{23.58} & 33.44 & \textbf{46.46} & 0.323 & \textbf{0.462} & 0.91 & \textbf{2.87} & 0.59 & \textbf{0.84} \\
 & Qwen 1.7B & 21.77 & \textbf{33.29} & 46.98 & \textbf{51.95} & 0.483 & \textbf{0.579} & 4.23 & \textbf{5.26} & 0.86 & \textbf{1.26} \\
 & Qwen 4B & 21.74 & \textbf{21.78} & 44.47 & \textbf{46.98} & \textbf{0.522} & 0.484 & 3.84 & \textbf{4.22} & 0.86 & \textbf{0.86} \\
 & Llama 1B & 8.88 & \textbf{13.81} & 37.38 & \textbf{43.12} & 0.368 & \textbf{0.454} & 2.86 & \textbf{3.68} & 0.41 & \textbf{0.73} \\
 & Llama 3B & 6.12 & \textbf{9.47} & 30.08 & \textbf{37.33} & 0.319 & \textbf{0.375} & 2.13 & \textbf{2.95} & 0.26 & \textbf{0.47} \\
\bottomrule
\end{tabular}%
}
\vspace{0.1cm}
\caption{Comparison of standard DP-LoRA vs.\ our denoising method on E2E and DART generation datasets after 50 steps. Bold indicates the better value within each pair.}
\label{tab:e2e_dart_results_new_50}
\end{table}

\section{Conclusion}

In this work, we identified the noise-induced degradation of gradient singular-value structure as a critical bottleneck in differentially private LLM fine-tuning. Using random matrix theory, we first explained why this degradation—specifically, a singular-value blow-up—happens under DP-SGD. Second, we formulated a method to test whether this degradation actually reduces performance. Our results conclusively show that it does. While other denoising methods such as DOPPLER and DiSK exploit temporal information from a sequence of gradients across training steps, our method instead denoises each noisy batch gradient independently. By correcting the singular-value structure of each batch gradient, we demonstrated that restoring the singular values speeds up the optimization process without compromising privacy guarantees.
  
Across a diverse set of tasks—from GLUE classification benchmarks with RoBERTa to generation tasks with multi-billion parameter Qwen and Llama models—the method with the corrected singular-value structure consistently accelerated convergence. We achieved training speedups ranging from 15.6\% to 75.0\% compared to the standard DP-LoRA baseline. Importantly, this accelerated convergence means that our approach reaches the exact same accuracy while saving significant privacy budget. 

We believe this work opens a new research direction for privacy-preserving machine learning. Future methods for low-rank DP fine-tuning should intrinsically maintain the singular-value structure (i.e., fast decay) of the gradients, which can ultimately lead to substantially improved privacy guarantees and convergence speed.

\section*{Software and Data}

All code and hyperparameters needed to reproduce the results and figures in this paper are available at \url{https://github.com/alidadsetan/re-exam-private-low-rank}.

\section*{Impact Statement}

This work addresses the critical challenge of efficient differentially private training for large language models, a key step towards mitigating privacy risks in natural language processing applications. By improving the trade-off between privacy and utility, our method facilitates the responsible deployment of models trained on sensitive data. However, we note that the strong guarantees of differential privacy rely heavily on correct implementation and strict adherence to the formalism, which we do not guarantee for the code provided as supplementary material. Additionally, these privacy guarantees are not absolute; there remains a non-zero privacy loss, quantified by the privacy budget. Furthermore, prior research has shown that DP-SGD can exacerbate fairness gaps, disproportionately affecting underrepresented groups. While our approach improves overall convergence speed, practitioners should carefully evaluate and monitor the fairness implications of deploying these models in real-world settings.

\nocite{langley00}

\begin{ack}
Ali Dadsetan acknowledges partial support from the Abdul Majid Bader Graduate Scholarship, awarded by Dalhousie University. Frank Rudzicz acknowledges support from the Canada CIFAR AI Chairs program.
Resources used in preparing this research were provided, in part, by the Province of Ontario, the Government of Canada through CIFAR, and companies sponsoring the Vector Institute (\url{https://www.vectorinstitute.ai/partnerships/current-partners/}), and the Digital Research Alliance of Canada (\url{https://alliancecan.ca}).
The authors thank the anonymous reviewers for their constructive feedback, which helped improve the manuscript.
\end{ack}

\medskip
\bibliography{example_paper}
\bibliographystyle{plainnat}

%%%%%%%%%%%%%%%%%%%%%%%%%%%%%%%%%%%%%%%%%%%%%%%%%%%%%%%%%%%%
\clearpage
\appendix

\raggedbottom
\section{Singular Value Distribution of Gradients Under DP Noise}
\label{appendix:sv_distribution}

\begin{figure}[h]
   \centering
   \begin{subfigure}[t]{0.48\linewidth}
      \centering
      \includegraphics[width=\linewidth]{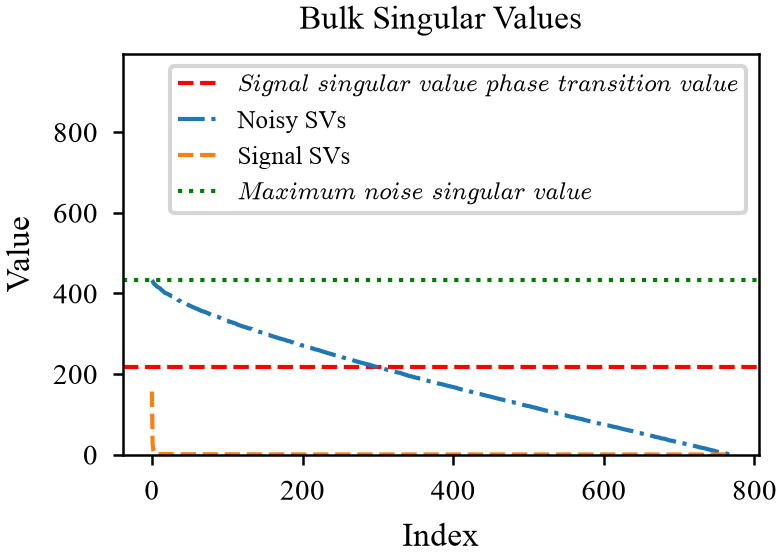}
      \caption{Indistinguishable from noise}
      \label{fig:roberta_svs_bulk}
   \end{subfigure}
   \hfill
   \begin{subfigure}[t]{0.48\linewidth}
      \centering
      \includegraphics[width=\linewidth]{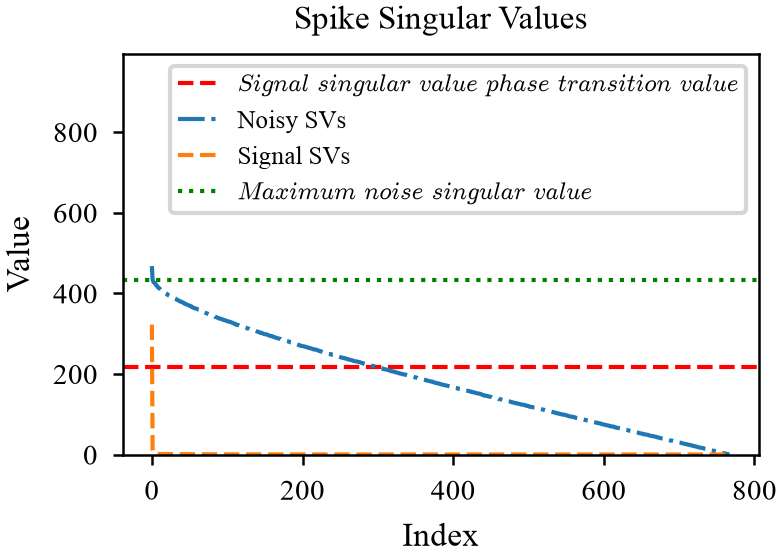}
      \caption{Deviating from the bulk}
      \label{fig:roberta_svs_spike}
   \end{subfigure}
   \caption{Sorted singular values of the gradient matrix for a RoBERTa layer, before and after adding DP-SGD noise. (a) When signal singular values are smaller than the red line, the noisy spectrum is indistinguishable from pure noise. (b) When some signal singular values exceed the red line, the largest singular values deviate from the bulk.}
   \label{fig:roberta_svs}
\end{figure}

The singular values of the gradient matrix undergo a ``phase transition'' \citep{baik2005phase} when noise is added. If the underlying signal is weak, the singular values of the noisy matrix become indistinguishable from those of pure noise. Figure~\ref{fig:roberta_svs_bulk} illustrates this by comparing the sorted singular values of a RoBERTa layer's gradient matrix before and after DP-SGD noise is applied. In this weak-signal regime, the noisy gradient's singular values closely follow the ``bulk'' distribution predicted by the Marchenko--Pastur law \citep{marvcenko1967distribution,tao2012topics}, making them essentially indistinguishable from pure noise. Thus, when a low-rank signal is too small relative to the noise, it is hidden in the noise and cannot be detected or recovered by examining the singular values and vectors alone. This highlights a fundamental limitation: sufficiently weak signals are undetectable in the presence of strong noise.

However, if some singular values exceed this threshold, the largest singular values of the noisy matrix deviate from the bulk, as shown in Figure~\ref{fig:roberta_svs_spike}. This phenomenon is known as the Baik--Ben Arous--P\'ech\'e (BBP) phase transition \citep{baik2005phase}. The extent of these deviations, as well as the alignment between the singular vectors of the noisy and original matrices, can be predicted mathematically \citep{baik2006eigenvalues,benaych2012singular}. These properties enable partial recovery of the original matrix from its noisy observation \citep{shabalin2013reconstruction,gavish2014optimal}.

\clearpage
\section{Finite-dimensional derivation of random matrix theory results}
\label{appendix:rmt_derivation}
In the random matrix theory literature, the results in \cite{shabalin2013reconstruction,donoho2018optimal,gavish2014optimal} are stated in the asymptotic regime, where the matrix dimensions grow to infinity and the noise variance may scale with the dimensions. In this section, we state those results in their original form and explain the derivation of equations~\ref{eq:sing_val}, \ref{eq:sim_u}, and \ref{eq:sim_v} from their asymptotic forms.

The setup in \cite{shabalin2013reconstruction,donoho2018optimal} is as follows. We have a sequence of matrices \(\bm{X}_n \in \mathbb{R}^{m_n \times n}\) with \(m_n/n \to \beta\) as \(n \to \infty\). The rank of the signal matrix is fixed, i.e. \(\mathrm{rank}(\bm{X}_n) = r\) for all \(n\). The singular values of the signal matrix are fixed, i.e. the non-zero singular values of \(\bm{X}_n\) are \(\lambda_1 > \lambda_2 > \ldots > \lambda_r > 0\) for all \(n\). Then, we add a noise matrix with i.i.d. entries from \(\mathcal{N}(0, 1/n)\) to get the noisy matrix. In these settings, the results in \cite{shabalin2013reconstruction,gavish2014optimal} state that
\begin{equation}
   \label{eq:asymp_sing_val}
\lim_{n \to \infty} y_{n,i} \overset{a.s.}{=} 
\begin{cases}
\displaystyle
\sqrt{\left(\lambda_i + \frac{1}{\lambda_i}\right)\left(\lambda_i + \frac{\beta}{\lambda_i}\right)} & \lambda_i > \beta^{1/4} \\[12pt]
1 + \sqrt{\beta} & \lambda_i \leq \beta^{1/4}
\end{cases}
\end{equation}
where \(y_{n,i}\) is the \(i\)-th singular value of the noisy matrix. To obtain the finite-dimensional form, we start from a noise matrix with i.i.d. entries from \(\mathcal{N}(0, \sigma^2)\) instead of \(\mathcal{N}(0, 1/n)\). Rescaling the observation as \(\frac{Y}{\sigma\sqrt{n}}\) gives the noise entries the desired distribution. Applying equation~\ref{eq:asymp_sing_val} to \(\frac{Y}{\sigma\sqrt{n}}\) and substituting \(\beta = m/n\) yields equation~\ref{eq:sing_val}. Similar arguments yield equations~\ref{eq:sim_u} and~\ref{eq:sim_v} from their asymptotic forms in \cite{shabalin2013reconstruction}.
\clearpage
\section{Matrix Denoising Details}
\label{appendix:matrix_denoising} 

\paragraph{Optimal Denoising}
The optimal estimator for the signal matrix can be written as

\begin{equation}
\label{eq:optimal_denoising}
\hat{\bm{X}}_{\mathrm{optimal}} = \sum_{i=1}^{r} \eta_i \tilde{\bm{u}}_i \tilde{\bm{v}}_i^T
\end{equation}

where, following \citet{shabalin2013reconstruction}, the optimal coefficients are

\begin{equation}
\label{eq:optimal_eta}
\eta_i = \hat{\lambda}_i \cdot \sqrt{\frac{\hat{\lambda}_i^4 - mn\sigma^4}{\hat{\lambda}_i^4 + m\hat{\lambda}_i^2\sigma^2}} \cdot \sqrt{\frac{\hat{\lambda}_i^4 - mn\sigma^4}{\hat{\lambda}_i^4 + n\hat{\lambda}_i^2\sigma^2}},
\end{equation}
where
\begin{equation*}
\hat{\lambda}_i = F^{-1}_{\sigma, n, m}(\tilde{\lambda}_i)
\end{equation*}

for all \(i\) such that \(\tilde{\lambda}_i > \sigma(\sqrt{m} + \sqrt{n})\), and zero otherwise. It has been shown to achieve the best possible mean squared error (MSE) under certain conditions, particularly when the noise is Gaussian and the signal is low-rank.

\paragraph{Computational Overhead}
\label{par:computational_overhead}
Although matrix denoising requires additional computation, careful implementation and parallelization can keep the overhead minimal. Three properties help reduce this overhead. First, in equation~\ref{eq:optimal_denoising}, \(F^{-1}_{\sigma, n, m}(\tilde{\lambda}_i)\) can be computed independently for each singular value, as can \(\eta_i\). Second, the SVD, which is the most computationally expensive part of the algorithm, can be computed in batches for network layers with similar dimensions. Third, approximate methods can further accelerate the SVD when acceptable. A na\"ive implementation can incur overhead of just under \(3\%\) of total training time; combining these strategies reduces the overhead to less than \(1\%\) in our experiments (Table~\ref{tab:computational_overhead}). The efficient implementation is available at \url{https://github.com/alidadsetan/re-exam-private-low-rank}.

\begin{table}[h]
\centering
\caption{Training time in seconds for different methods of DP-SGD fine-tuning of RoBERTa on SST-2, using the settings in Section~\ref{sec:experiments}. Overhead is the percentage increase in training time compared with regular DP-SGD.}
\label{tab:computational_overhead}
\resizebox{0.9\columnwidth}{!}{%
\begin{tabular}{l l r r}
\toprule
\textbf{Model} & \textbf{Method} & \textbf{Train Time} & \textbf{Overhead} \\
\midrule

\multirow{3}{*}{RoBERTa Base} 
  & Regular DP-SGD                   & 1846 & --    \\
  & Na\"ive denoising implementation & 1899 & 2.87\% \\
  & Efficient denoising implementation & 1861 & 0.81\% \\
\midrule

\multirow{3}{*}{RoBERTa Large} 
  & Regular DP-SGD                   & 5658 & --    \\
  & Na\"ive denoising implementation & 5781 & 2.17\% \\
  & Efficient denoising implementation & 5700 & 0.74\% \\
\bottomrule
\end{tabular}%
}
\end{table}

\subsection{Why optimal denoising improves cosine similarity}
\label{appendix:why_optimal_denoising}

The MSE optimality of the denoiser in equation~\ref{eq:optimal_denoising} is an asymptotic result: under the low-rank signal-plus-Gaussian-noise model, it minimizes the limiting squared Frobenius error within the considered class of singular-value estimators \citep{shabalin2013reconstruction,gavish2017optimal}. We use this almost-sure limiting loss optimality to show that the denoiser also maximizes limiting cosine similarity within the same class, and that norm correction preserves this alignment.

Consider the normalized model from Appendix~\ref{appendix:rmt_derivation},
\[
\bm{Y}_n=\bm{X}_n+\bm{Z}_n,
\qquad \bm{X}_n\in\mathbb{R}^{m_n\times n},
\qquad \frac{m_n}{n}\longrightarrow\beta\in(0,1],
\]
where the deterministic signal has fixed rank and fixed nonzero singular values, and the entries of \(\bm{Z}_n\) are independent \(\mathcal{N}(0,1/n)\) variables. Let \(\mathcal{E}\) be the considered class of admissible singular-value shrinkage rules, with estimators
\[
\widehat{\bm{X}}_{\eta,n}
=\sum_{i=1}^{\min(m_n,n)}\eta(y_{n,i})\,\tilde{\bm{u}}_{n,i}\tilde{\bm{v}}_{n,i}^{\top},
\qquad \eta\in\mathcal{E},
\]
where \(y_{n,i},\tilde{\bm{u}}_{n,i},\tilde{\bm{v}}_{n,i}\) are the singular values and vectors of \(\bm{Y}_n\). For nonzero matrices, write
\[
\cos(\bm{A},\bm{B})=
\frac{\langle\bm{A},\bm{B}\rangle_F}{\|\bm{A}\|_F\|\bm{B}\|_F},
\qquad \langle\bm{A},\bm{B}\rangle_F=\operatorname{tr}(\bm{A}^{\top}\bm{B}).
\]
The identity
\begin{equation}
\label{eq:frobenius_cosine_identity}
\|\bm{A}-\bm{B}\|_F^2
=\|\bm{A}\|_F^2+\|\bm{B}\|_F^2
-2\|\bm{A}\|_F\|\bm{B}\|_F\cos(\bm{A},\bm{B})
\end{equation}
shows that, for a fixed target and equal estimator norms, minimizing squared error is equivalent to maximizing cosine similarity. The following theorem extends this observation to estimators with different norms by matching their limiting norms.

\begin{theorem}[Asymptotic alignment optimality]
\label{theorem:asymptotic_alignment}
Suppose \(\mathcal{E}\) is closed under positive constant scaling: \(\eta\in\mathcal{E}\) implies \(c\eta\in\mathcal{E}\) for every \(c>0\). Let \(\widehat{\bm{X}}_{F,n}\) be asymptotically Frobenius-optimal in this class, so that, almost surely,
\[
\lim_{n\to\infty}\|\bm{X}_n-\widehat{\bm{X}}_{F,n}\|_F^2
\leq
\lim_{n\to\infty}\|\bm{X}_n-\widehat{\bm{X}}_{\eta,n}\|_F^2
\quad\text{for every admissible }\eta
\]
whenever these limits exist. Assume the following almost-sure norm limits are deterministic, finite, and strictly positive:
\[
s=\lim_{n\to\infty}\|\bm{X}_n\|_F,
\qquad r_F=\lim_{n\to\infty}\|\widehat{\bm{X}}_{F,n}\|_F,
\qquad r_\eta=\lim_{n\to\infty}\|\widehat{\bm{X}}_{\eta,n}\|_F.
\]
For every \(\eta\in\mathcal{E}\) satisfying these conditions and for which both cosine limits exist, we have, almost surely,
\begin{equation}
\label{eq:asymptotic_alignment}
\lim_{n\to\infty}\cos(\widehat{\bm{X}}_{F,n},\bm{X}_n)
\geq
\lim_{n\to\infty}\cos(\widehat{\bm{X}}_{\eta,n},\bm{X}_n).
\end{equation}
\end{theorem}

\begin{proof}
Fix an admissible \(\eta\) and denote the two cosine limits by \(C_F\) and \(C_\eta\). Set \(c=r_F/r_\eta>0\). Closure under scaling makes \(c\eta\) admissible, and
\[
\widehat{\bm{X}}_{c\eta,n}=c\widehat{\bm{X}}_{\eta,n},
\qquad
\lim_{n\to\infty}\|\widehat{\bm{X}}_{c\eta,n}\|_F=cr_\eta=r_F.
\]
Positive scaling preserves cosine similarity. Applying equation~\ref{eq:frobenius_cosine_identity} and taking limits therefore gives
\begin{align*}
\lim_{n\to\infty}\|\bm{X}_n-\widehat{\bm{X}}_{F,n}\|_F^2
&=s^2+r_F^2-2sr_FC_F,\\
\lim_{n\to\infty}\|\bm{X}_n-\widehat{\bm{X}}_{c\eta,n}\|_F^2
&=s^2+r_F^2-2sr_FC_\eta.
\end{align*}
Frobenius optimality makes the first expression no larger than the second. Since \(s,r_F>0\), this yields \(C_F\geq C_\eta\).
\end{proof}

\paragraph{Norm correction preserves alignment.}
For any prescribed norm \(R_n>0\), including the norm of the noisy layer gradient, define
\[
\overline{\bm{X}}_{\eta,n}
=R_n\frac{\widehat{\bm{X}}_{\eta,n}}{\|\widehat{\bm{X}}_{\eta,n}\|_F}.
\]
Whenever the estimator is nonzero,
\[
\cos(\overline{\bm{X}}_{\eta,n},\bm{X}_n)
=\cos(\widehat{\bm{X}}_{\eta,n},\bm{X}_n).
\]
Thus norm correction preserves the asymptotic cosine-similarity ordering in equation~\ref{eq:asymptotic_alignment}. This is an alignment guarantee; rescaling need not preserve Frobenius-loss optimality. Matching each output to its original noisy layer norm then allows per-layer alignment improvements to transfer to the full gradient by Theorem~\ref{theorem:norm_correction}.

\paragraph{Relation to noisy gradients and finite-dimensional layers.}
In the normalized fixed-rank model above, \(\|\bm{Y}_n\|_F\to\infty\) while \(\langle\bm{Y}_n,\bm{X}_n\rangle_F\to s^2\) almost surely, so \(\cos(\bm{Y}_n,\bm{X}_n)\to0\). The noisy matrix therefore lies outside the finite-norm estimator class in the theorem. For a Frobenius-optimal estimate with positive limiting norm \(r_F>0\), optimality under rescaling requires \(a=1\) to minimize \(s^2+a^2r_F^2-2asr_FC_F\) over \(a>0\), giving \(C_F=r_F/s>0\). This separately establishes an asymptotic alignment improvement over the noisy matrix when a nonzero signal estimate survives.

These statements concern the limit \(n\to\infty\), and do not guarantee improvement for every finite-dimensional layer. When all signal components lie below the recovery threshold, \(r_F=0\) and the theorem does not apply; whenever the denoiser returns the zero matrix, its cosine similarity is undefined. Our choice \(\kappa=1.02\) supplies an empirical margin above the asymptotic noise edge before applying denoising, as discussed in Appendix~\ref{sec:why_threshold}; layers below this threshold retain their noisy gradients. The limiting theorem thus motivates denoising, while the threshold addresses its finite-dimensional application.

\clearpage
\section{Additional results on noise-induced slowdown}
\label{appendix:clip_vs_dp}
\begin{figure}[h]
   \centering
   \includegraphics[width=0.7\linewidth]{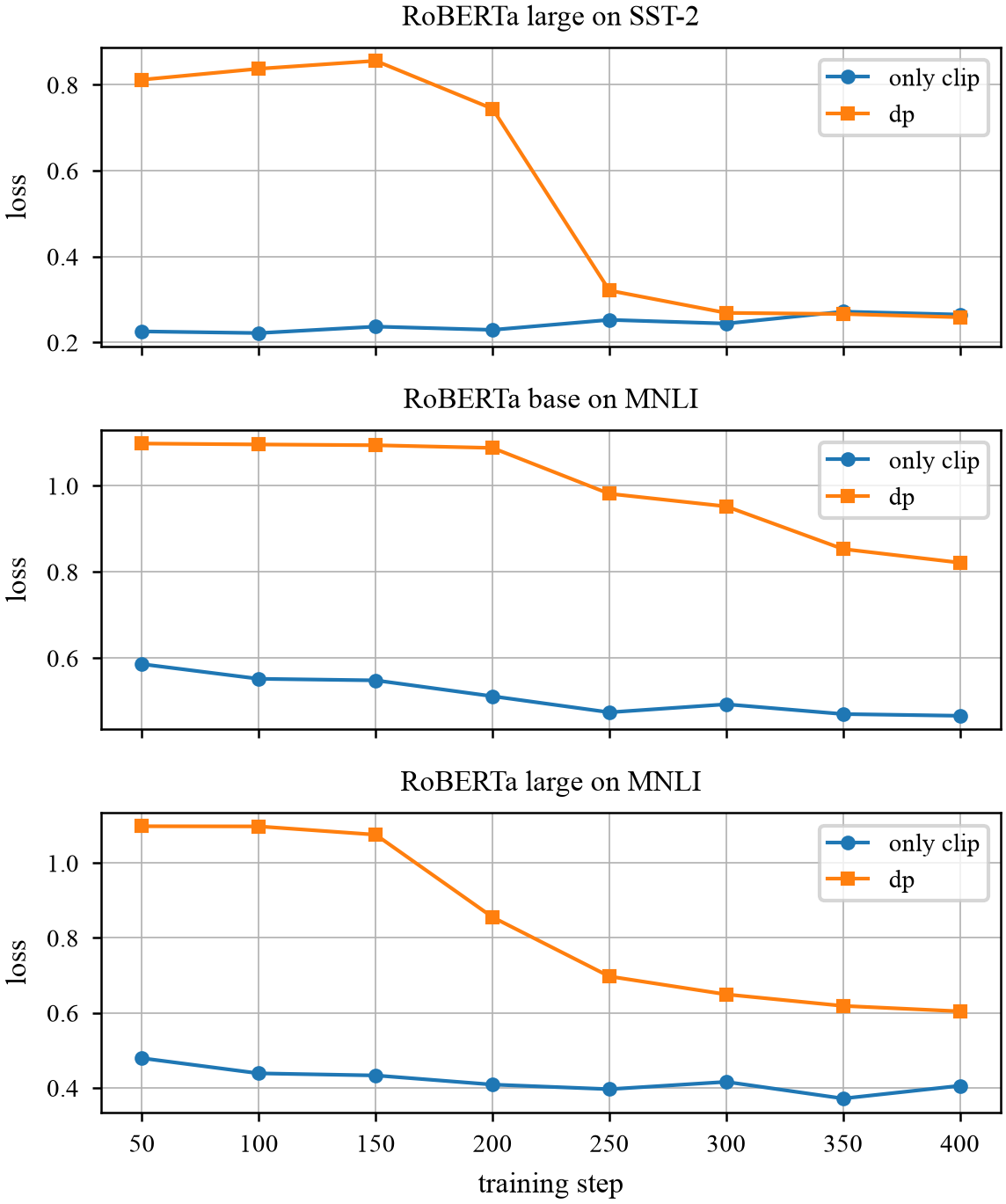}
   \caption{Validation loss curves for DP-SGD with and without noise, using RoBERTa models on SST-2 and MNLI. Adding noise slows convergence compared with clipping-only training. For some model/dataset pairs, a substantial gap also remains in the final loss.}
   \label{fig:clip_vs_dp_additional}
\end{figure}

As discussed in the introductory paragraph of Section~\ref{sec:slow-down}, two factors can slow DP-SGD convergence: per-sample gradient clipping and noise addition. Figure~\ref{fig:clip_vs_dp_additional} provides additional empirical evidence of the effect of noise.

\clearpage
\section{Denoising Function Details}
\label{appendix:denoising_details}

This appendix provides the full specification of the denoising function and justifications for its two key design choices.

\subsection{Full Denoising Function Specification}
\label{appendix:full_denoising_spec}
Our denoising function applies the method in Appendix~\ref{appendix:matrix_denoising} to the linear-layer components of the noisy gradient \(\tilde{\bm{g}}\). For a layer \(\bm{W}\) of our neural network \(\theta\), the restriction of the clipped gradient to \(\bm{W}\) is the matrix \(\sum_{x\in\mathcal{B}}\mathrm{clip}(\nabla_{\theta} f(\theta, x), C)\!\mid_{\bm{W}} = \bar{\bm{g}}\!\mid_{\bm{W}}\). The network parameters can be partitioned by layer as
\[
\theta = \bm{W}_1 \times \bm{W}_2 \times \ldots \times \bm{W}_L
\]
where \(L\) is the number of layers in the network. Then, we can write 

\begin{align*}
\bar{\bm{g}} &= (\bar{\bm{g}}\!\mid_{W_1}, \bar{\bm{g}}\!\mid_{W_2}, \ldots, \bar{\bm{g}}\!\mid_{W_L}) \\
\tilde{\bm{g}} &= (\tilde{\bm{g}}\!\mid_{W_1}, \tilde{\bm{g}}\!\mid_{W_2}, \ldots, \tilde{\bm{g}}\!\mid_{W_L})
\end{align*}

With this notation, we define the full denoising function by applying the per-layer denoising function separately to each layer's gradient:

\[
\mathrm{Denoise}(\tilde{\bm{g}}) = \left(\mathrm{Denoise}(\tilde{\bm{g}}\!\mid_{W_1}),  \ldots, \mathrm{Denoise}(\tilde{\bm{g}}\!\mid_{W_L})\right)
\]

If a layer \(\bm{W}_i\) is not linear, we simply set \(\mathrm{Denoise}(\tilde{\bm{g}}_{W_i}) = \tilde{\bm{g}}_{W_i}\). We also define the per-layer improvement as
\begin{align*}
\mathrm{Improvement}_{i}(t) &= \mathrm{cos}(\mathrm{Denoise}(\tilde{\bm{g}}_t\!\mid_{\bm{W}_i}), \bar{\bm{g}}_t\!\mid_{\bm{W}_i}) \\
&\quad - \mathrm{cos}(\tilde{\bm{g}}_t\!\mid_{\bm{W}_i}, \bar{\bm{g}}_t\!\mid_{\bm{W}_i}).
\end{align*}

For linear layers, we modify the optimal denoising method in two ways:
\begin{itemize}
   \item \textbf{To adapt the asymptotic formulas to finite dimensions}, we apply optimal denoising only if the largest singular value of the noisy layer gradient exceeds a preset multiple of \(\sigma(\sqrt{n} + \sqrt{m})\), where \(n, m\) are the dimensions of that layer.
   \item \textbf{To preserve gradient norms and ensure that per-layer alignment improvements transfer to the full gradient}, we rescale each denoised layer gradient to match the \(\ell_2\) norm of its noisy counterpart: \( \tilde{\bm{g}}|_{\bm{W}}\mapsto \frac{||\tilde{\bm{g}}|_{\bm{W}}||}{||\hat{\bm{g}}_{\mathrm{optimal}}||} \hat{\bm{g}}_{\mathrm{optimal}}\), where \(\hat{\bm{g}}_{\mathrm{optimal}}\) is the optimal denoising of \(\tilde{\bm{g}}|_{\bm{W}}\) (equation~\ref{eq:optimal_denoising}).
\end{itemize} 
Thus, for linear layers and threshold hyperparameter \(\kappa\), we have
\begin{equation}
\label{eq:denoising_function}
\mathrm{Denoise}(\tilde{\bm{g}}|_{\bm{W}}) = 
\begin{cases}
    \tilde{\bm{g}}|_{\bm{W}} & \lambda_1 < \tau \\
    \frac{\|\tilde{\bm{g}}|_{\bm{W}}\|}{\|\hat{\bm{g}}_{\mathrm{optimal}}\|} \hat{\bm{g}}_{\mathrm{optimal}} & \text{otherwise}
\end{cases}
\end{equation}
where \(\tau = \kappa\sigma(\sqrt{n} + \sqrt{m})\) and \(\lambda_1 = \lambda_1(\tilde{\bm{g}}|_{\bm{W}})\). The noise standard deviation \(\sigma\) differs from the DP-SGD noise multiplier; the two are related by
\[
\sigma = \frac{\sigma_{\mathrm{DP}}C}{|\mathcal{B}|}.
\]
\subsection{Why is a threshold needed, and why this value?}
\label{sec:why_threshold}
The results in Section~\ref{sec:matrix_denoising} are derived in an asymptotic setting. For instance, the theory predicts that if all singular values of the signal matrix are less than \(\sigma\sqrt[4]{nm}\), or equivalently, if all singular values of the noisy matrix are less than \(\sigma(\sqrt{n} + \sqrt{m})\), then the inner products between the left (or right) singular vectors of the signal and noisy matrices should be zero. In that case, the optimal denoising algorithm returns the zero matrix as the best estimate. For finite-dimensional matrices, however, the noisy gradient usually retains some positive cosine similarity with the original gradient, even when its singular values are small. As a result, denoising does not always improve alignment with the per-layer clipped gradient.

Fortunately, we identified a simple RMT-based criterion to decide when to apply the denoiser. Concretely, we only denoise a layer if the largest singular value of its noisy gradient exceeds \(\kappa\,\sigma(\sqrt{n} + \sqrt{m})\). Choosing \(\kappa=1\) prevents denoising in cases where the optimal estimator would return the zero matrix. This threshold is motivated by the phase transition in equation~\ref{eq:sing_val}: \(\sigma(\sqrt{n} + \sqrt{m})\) is the maximal singular value of a pure-noise matrix (signal equal to zero). Thus, RMT indicates when denoising can meaningfully improve alignment; if the estimator would output the zero matrix, the theoretical prediction is that there is no signal to recover.

\begin{figure}[htbp]
   \centering
   \begin{subfigure}[t]{0.48\linewidth}
      \centering
      \includegraphics[width=\linewidth]{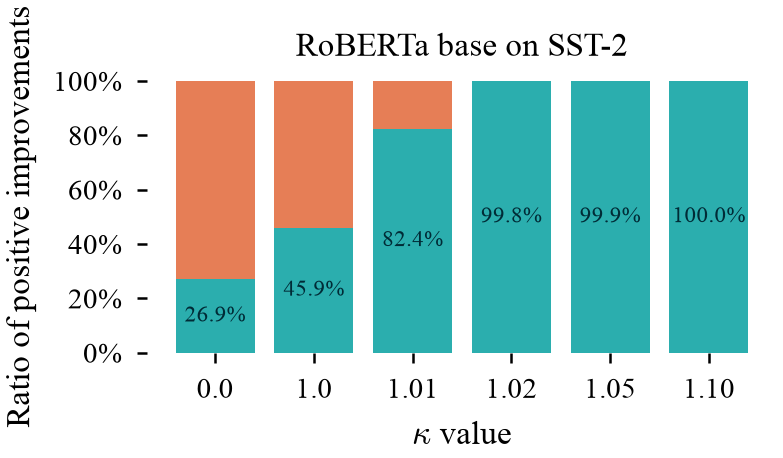}
      \caption{RoBERTa-base on SST-2}
      \label{fig:kappa_sweep_base_sst}
   \end{subfigure}
   \hfill
   \begin{subfigure}[t]{0.48\linewidth}
      \centering
      \includegraphics[width=\linewidth]{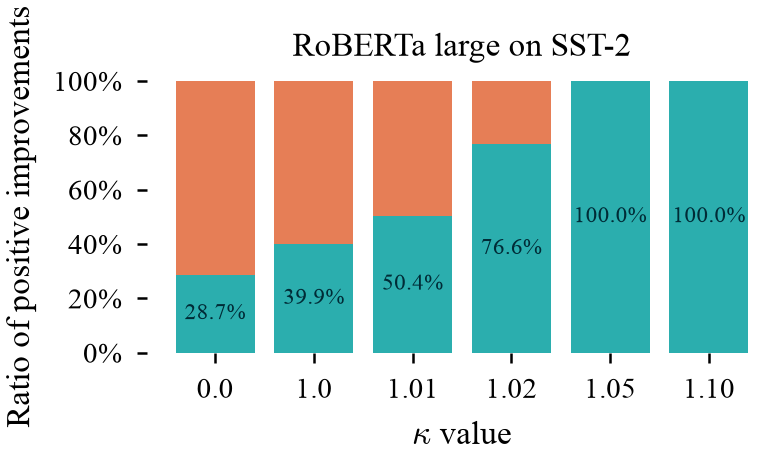}
      \caption{RoBERTa-large on SST-2}
      \label{fig:kappa_sweep_large_sst}
   \end{subfigure}

   \medskip
   \begin{subfigure}[t]{0.48\linewidth}
      \centering
      \includegraphics[width=\linewidth]{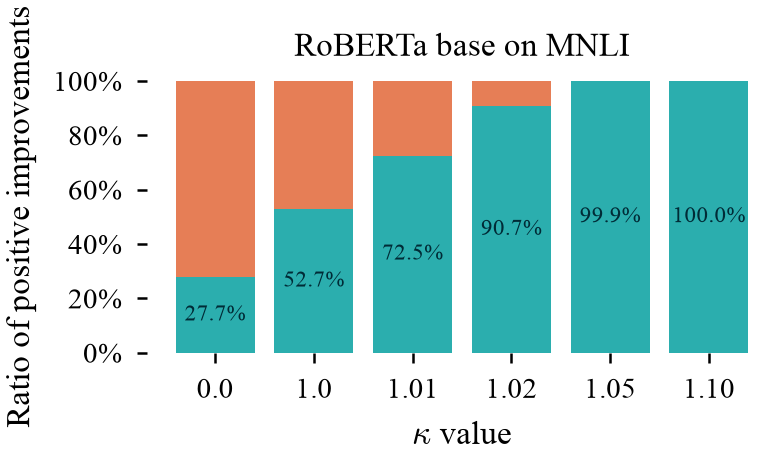}
      \caption{RoBERTa-base on MNLI}
      \label{fig:kappa_sweep_base_mnli}
   \end{subfigure}
   \hfill
   \begin{subfigure}[t]{0.48\linewidth}
      \centering
      \includegraphics[width=\linewidth]{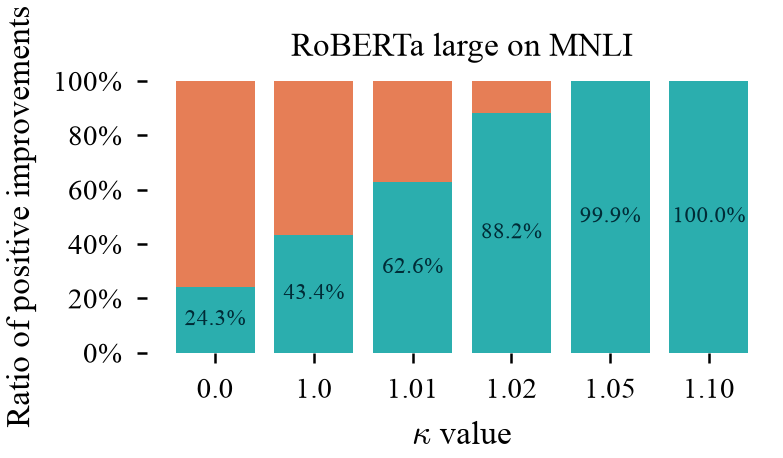}
      \caption{RoBERTa-large on MNLI}
      \label{fig:kappa_sweep_large_mnli}
   \end{subfigure}
   \caption{Effect of \(\kappa\) on per-layer alignment improvement for RoBERTa-base and RoBERTa-large on SST-2 and MNLI. The vertical axis gives the percentage of layers whose cosine similarity with the clipped gradient improves after denoising, among layers whose largest singular value exceeds \(\kappa\sigma(\sqrt{n}+\sqrt{m})\). Settings follow Section~\ref{sec:experiments}. For RoBERTa-base on SST-2, \(\kappa=1.02\) is the smallest tested value yielding predominantly positive per-layer improvements.}
   \label{fig:kappa_sweeps}
\end{figure}

Our observations show that a suitable value of \(\kappa\) tends to improve alignment; otherwise, denoising may reduce it. We tuned \(\kappa\) on SST-2 with RoBERTa-base by choosing the best value from \(\{1.01, 1.02, 1.05, 1.1\}\), then used the same value for all other model/dataset pairs. We chose values greater than \(1\) to avoid cases where denoising returns the zero matrix, while keeping them close to \(1\) so that more layers are denoised. The trade-off is to make \(\kappa\) large enough for denoising to improve alignment, but small enough to denoise enough layers to benefit the full gradient. The best value was \(\kappa = 1.02\) (Figure~\ref{fig:kappa_sweep_base_sst}). Although we could have tuned \(\kappa\) for each model/dataset pair, we used a common value to demonstrate robustness to this hyperparameter. The value \(\kappa = 1.02\) worked well across all tested model/dataset pairs.

\begin{figure}
   \centering
   \includegraphics[width=\linewidth]{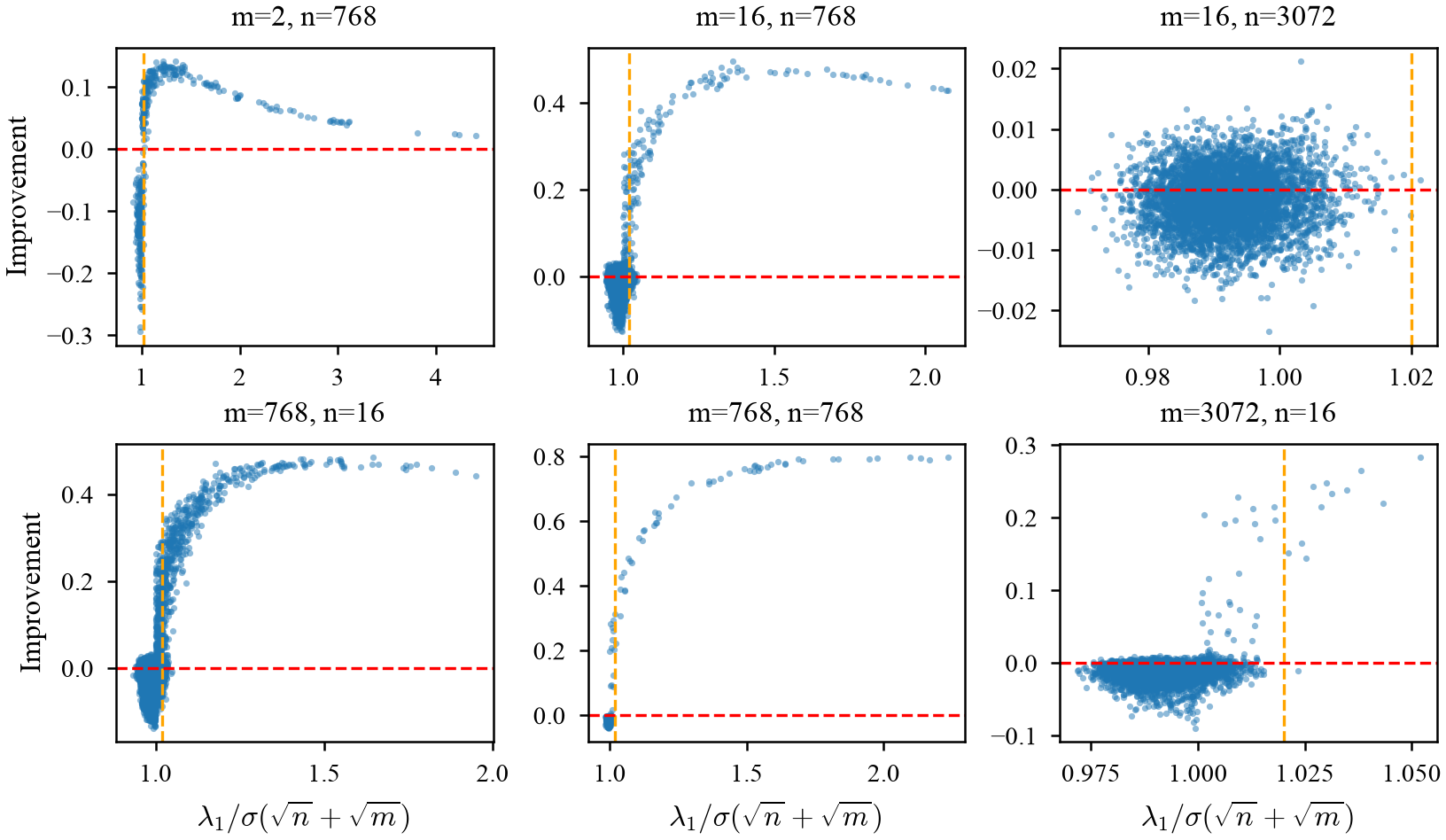}
   \caption{Per-layer alignment improvement versus \(\frac{\lambda_1}{\sigma(\sqrt{n} + \sqrt{m})}\) for different layer dimensions. The vertical yellow line marks the threshold \(\kappa\) used in our experiments. The aim is to include many points with positive improvement to its right, while excluding points with negative improvement and omitting as few positive-improvement points as possible to its left.}
   \label{fig:denoising_alignment}
\end{figure}

\begin{figure}
   \centering
   \includegraphics[width=0.9\linewidth]{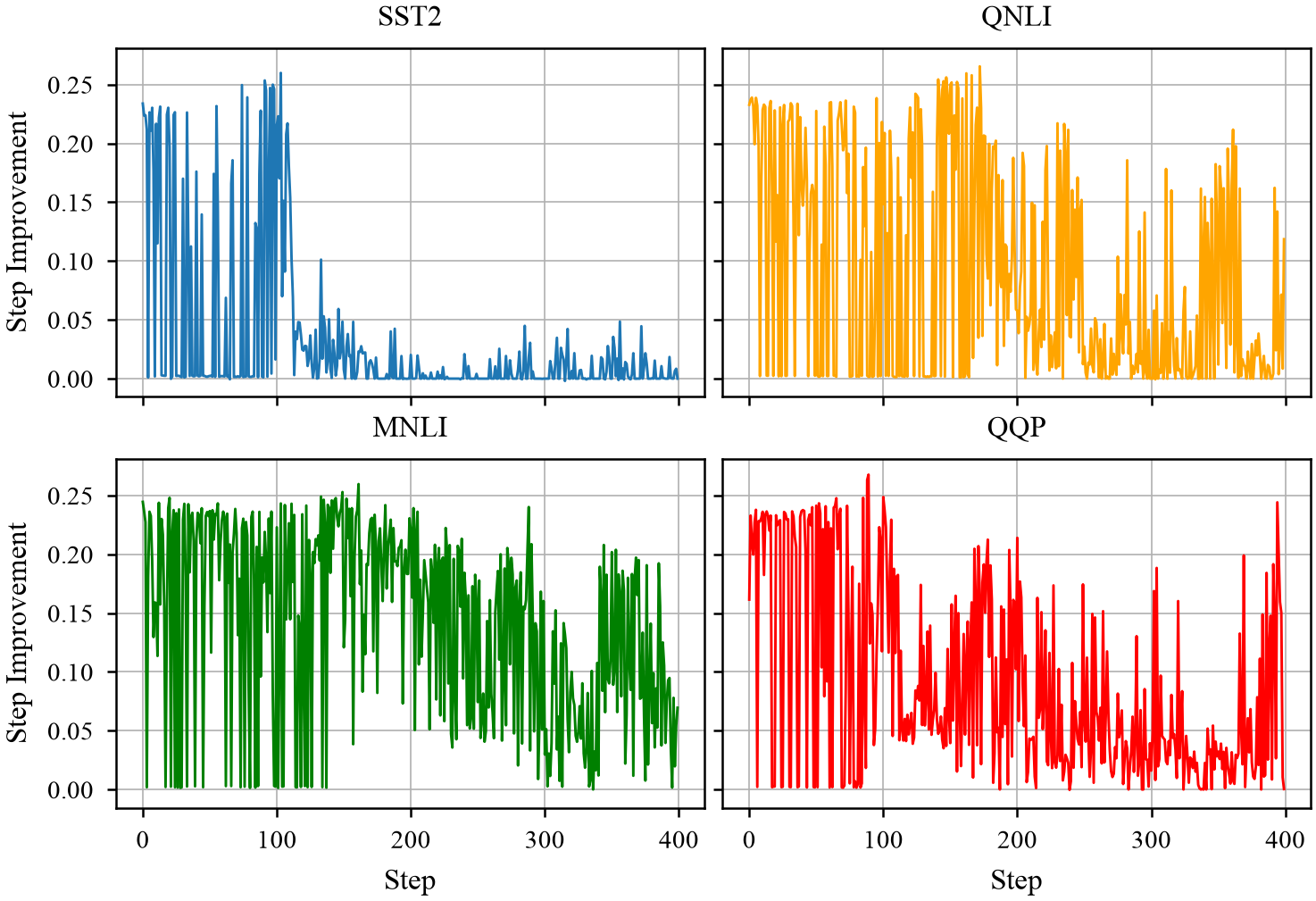}
   \caption{Cosine-similarity improvement from denoising, evaluated against the clipped gradient across training steps and datasets. Positive values indicate improved alignment after denoising.}
   \label{fig:improvements}
\end{figure}

\paragraph{Additional results.}
Additional hyperparameter sweeps for \(\kappa\) across model/dataset pairs are shown in the remaining panels of Figure~\ref{fig:kappa_sweeps}. Figure~\ref{fig:denoising_alignment} plots per-layer alignment improvement against \(\frac{\lambda_1}{\sigma(\sqrt{n} + \sqrt{m})}\) for different layer dimensions, further supporting our threshold choice. Figure~\ref{fig:improvements} also shows whole-gradient cosine-similarity improvements over training steps.

\clearpage

\clearpage

\subsection{Why is norm correction needed?}
\label{sec:why_norm_correction}
The RMT denoiser addresses alignment with the clipped gradient within each layer; it does not account for the relative scaling of different layers. To extend per-layer alignment improvements to the full gradient vector, we use the following theorem: improving the alignment of each block while preserving its norm also improves overall alignment.

\begin{theorem}
   \label{theorem:norm_correction}
   Let \(x=(x_1,\dots,x_c)\in \mathbb{R}^n\) be a target vector, with \(x_i \in \mathbb{R}^{n_i}\) and \(\sum_{i=1}^c n_i = n\). Let \(y=(y_1,\dots,y_c), y'=(y'_1,\dots,y'_c)\in \mathbb{R}^n\) be estimates of \(x\), with \(y_i,y'_i \in \mathbb{R}^{n_i}\). If
   \begin{enumerate}
      \item[(i)] \(\mathrm{cos}(y_i,x_i) \leq \mathrm{cos}(y'_i,x_i)\), for all \(i\in\{1,\dots,c\}\); and
      \item[(ii)] \(\|y_i\|=\|y'_i\|\), for all \(i\in\{1,\dots,c\}\),
   \end{enumerate}
    then \(\mathrm{cos}(y,x) \leq \mathrm{cos}(y',x)\).
\end{theorem}

\begin{proof}
Since the blocks are disjoint and $\|y_i\|=\|y'_i\|$ for all $i$, we can write
\[
\langle y,x\rangle \;=\; \sum_{i=1}^c \langle y_i, x_i\rangle
\qquad\text{and}\qquad
\|y\|^2 \;=\; \sum_{i=1}^c \|y_i\|^2 \;=\; \sum_{i=1}^c \|y'_i\|^2 \;=\; \|y'\|^2 .
\]
For each block, cosine similarity is
\[
\cos(y_i,x_i) = \frac{\langle y_i,x_i\rangle}{\|y_i\|\|x_i\|}.
\]
Assumption (i) and the norm equality (ii) imply
\[
\langle y_i,x_i\rangle 
= \|y_i\|\|x_i\|\, \cos(y_i,x_i)
\;\le\;
\|y'_i\|\|x_i\|\, \cos(y'_i,x_i)
= \langle y'_i,x_i\rangle .
\]
Summing over all $i$ gives $\langle y,x\rangle \le \langle y',x\rangle$.  
Using the equality of global norms, we obtain
\[
\cos(y,x)
= \frac{\langle y,x\rangle}{\|y\|\,\|x\|}
\;\le\;
\frac{\langle y',x\rangle}{\|y'\|\,\|x\|}
= \cos(y',x).
\]
Thus $\cos(y,x)\le \cos(y',x)$, as claimed.
\end{proof}

In our case, the target vector is the clipped gradient \(\bar{\bm{g}}\), and the two estimates are the noisy gradient \(\tilde{\bm{g}}\) and the denoised gradient \(\mathrm{Denoise}(\tilde{\bm{g}})\). This theorem is a general result about cosine similarity and vector norms; it requires no matrix or low-rank structure. Unlike the results in Section~\ref{sec:matrix_denoising}, it is exact and relies on neither asymptotic approximations nor probabilistic arguments.

Furthermore, in Appendix~\ref{appendix:adam_lower_bound}, we show via a simplified quadratic model that improving cosine similarity between the denoised and true gradients improves the worst-case lower bound on loss reduction when using Adam, a commonly used optimizer for language models. While the result is derived under simplifying assumptions, it captures the intuition for why our similarity-based denoising objective translates into faster convergence in practice.

\begin{figure}[h]
   \centering
   \includegraphics[width=0.9\linewidth]{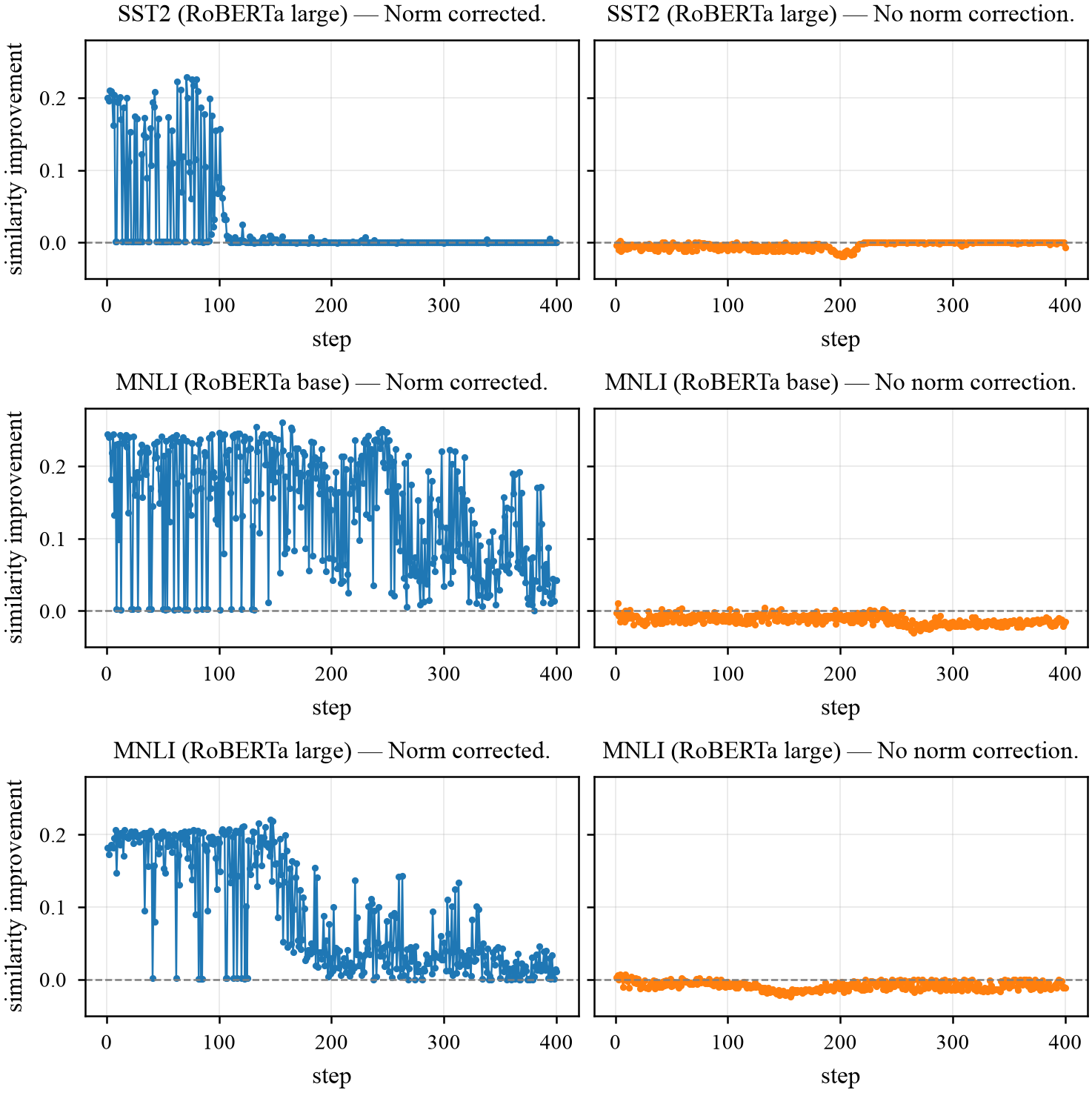}
   \caption{Comparison of whole-gradient cosine-similarity improvement with and without norm correction. Settings follow Section~\ref{sec:experiments}. Norm correction yields more consistent improvement.}
   \label{fig:norm_abal_additional}
\end{figure}

Assumption (ii) in Theorem~\ref{theorem:norm_correction} motivates the norm correction in equation~\ref{eq:denoising_function}. Without this correction, improved per-layer alignment does not guarantee improved overall alignment.
To evaluate the effect of norm correction, we conducted ablation studies. Figure~\ref{fig:norm_abal_sst_base} shows the results for RoBERTa-base on SST-2. Without norm correction, the improvement is inconsistent and sometimes negative, demonstrating the importance of this step. Additional ablation results appear in Figure~\ref{fig:norm_abal_additional}.

Finally, the denoised gradient has exactly the same norm as the noisy gradient. This prevents unexpected changes in gradient norm that could affect optimizer convergence.

\paragraph{Additional norm-correction ablation results.}
For completeness, the ablation replaces equation~\ref{eq:denoising_function} with the denoised gradient without norm correction:
\[
\mathrm{Denoise_{\textrm{unscaled}}}(\tilde{\bm{g}}\!\mid_{\bm{W}}) = 
\begin{cases}
    \tilde{\bm{g}}\!\mid_{\bm{W}}, & \text{if } \lambda_1(\tilde{\bm{g}}\!\mid_{\bm{W}}) < \kappa\,\sigma(\sqrt{n} + \sqrt{m}) \\[1.5ex]
   \hat{\bm{g}}_{\mathrm{optimal}}, & \text{otherwise}
\end{cases}
\]

In addition to Figure~\ref{fig:norm_abal_sst_base}, Figure~\ref{fig:norm_abal_additional} shows results for RoBERTa-base and RoBERTa-large on MNLI, and RoBERTa-large on SST-2. Without norm correction, the improvement is inconsistent and sometimes negative; with norm correction, it is consistently positive. These results further demonstrate the importance of norm correction.

\clearpage

\section{Alignment with the Unclipped Gradient}
\label{appendix:unclipped_alignment}
Our denoiser estimates the clipped gradient from its noisy observation, so improved alignment with this target does not by itself establish improved alignment with the gradient before clipping. To examine this distinction, we measure cosine-similarity improvement for RoBERTa-base on SST-2 and MNLI using both clipped and unclipped targets. For a target gradient \(\bm{h}_t\), we report
\[
\mathrm{Improvement}(t;\bm{h}_t)
=\cos\!\left(\mathrm{Denoise}(\tilde{\bm{g}}_t),\bm{h}_t\right)
-\cos\!\left(\tilde{\bm{g}}_t,\bm{h}_t\right),
\]
where \(\bm{h}_t\) is either the clipped gradient \(\bar{\bm{g}}_t\) or the unclipped gradient. Thus, positive values indicate that denoising improves alignment with the chosen target relative to the noisy gradient.

\begin{table}[htbp]
\centering
\caption{Cosine-similarity improvement from denoising for RoBERTa-base on SST-2 and MNLI, evaluated against clipped and unclipped gradients and averaged over the 50 training steps ending at each of eight checkpoints. Values are \(100\times\) the cosine-similarity differences defined above, reported in percent.}
\label{tab:unclipped_alignment}
\small
\setlength{\tabcolsep}{3.5pt}
\renewcommand{\arraystretch}{1.1}
\begin{tabular}{llrrrrrrrr}
\toprule
\textbf{Dataset} & \textbf{Target} & \textbf{Step 50} & \textbf{Step 100} & \textbf{Step 150} & \textbf{Step 200} & \textbf{Step 250} & \textbf{Step 300} & \textbf{Step 350} & \textbf{Step 400} \\
\midrule
\multirow{2}{*}{SST-2}
 & Clipped   & 9.38\% & 9.91\% & 5.08\% & 1.33\% & 1.17\% & 0.91\% & 1.15\% & 0.69\% \\
 & Unclipped & 6.31\% & 4.66\% & 3.54\% & 1.10\% & 0.94\% & 0.77\% & 0.98\% & 0.57\% \\
\midrule
\multirow{2}{*}{MNLI}
 & Clipped   & 16.74\% & 16.71\% & 16.98\% & 16.83\% & 9.49\% & 8.35\% & 7.77\% & 6.03\% \\
 & Unclipped & 16.31\% & 16.28\% & 16.29\% & 12.03\% & 6.87\% & 6.11\% & 5.50\% & 4.13\% \\
\bottomrule
\end{tabular}
\end{table}

Table~\ref{tab:unclipped_alignment} shows that improvements with respect to the unclipped gradient are smaller than those with respect to the clipped gradient at every reported checkpoint, but remain positive throughout. These results show that, despite clipping bias, our method also improves alignment with the unclipped gradient in these experiments.

\clearpage
\section{Quadratic Model: Improvement in Similarity Improves Worst-Case Loss Improvement in Adam}
\label{appendix:adam_lower_bound}

We study how the quality of the update direction used by Adam affects one-step progress. Our goal is to formalize the statement that, all else being equal, an update direction with larger cosine similarity to the true gradient gives better optimization improvement.

\subsection{Setup}

Let
\[
f(x) = \frac{1}{2}(x-x^\star)^\top H (x-x^\star), \quad H \succ 0,
\]
be a quadratic objective. At iterate \(x_t\), denote the true gradient by
\[
g_t := \nabla f(x_t) = H(x_t - x^\star).
\]

We model one Adam step as
\[
x_{t+1} = x_t - \alpha_t P_t s_t,
\]
where:
\begin{itemize}
    \item \(\alpha_t > 0\) is the step size,
    \item \(s_t \in \mathbb{R}^d\) is the vector used in the numerator of Adam's update (for example, a raw gradient estimate, a momentum-averaged estimate, or a denoised estimate),
    \item \(P_t \succ 0\) is Adam's diagonal preconditioner, \(P_t := \mathrm{diag}\left(\frac{1}{\sqrt{\hat{v}_t} + \varepsilon}\right)\).
\end{itemize}

For the purpose of a one-step comparison, we treat \(P_t\) as fixed. This isolates the effect of changing the update direction \(s_t\), while keeping the second-moment preconditioning unchanged.

We also assume that \(P_t\) is uniformly bounded:
\[
p_{\min} I \preceq P_t \preceq p_{\max} I
\]
for some constants \(0 < p_{\min} \le p_{\max} < \infty\).

Finally, for \(u, v \ne 0\), define the ordinary Euclidean cosine similarity by
\[
\cos(u,v) := \frac{u^\top v}{\|u\| \, \|v\|}.
\]

\subsection{Result 1: A one-step lower bound in terms of ordinary cosine}

\begin{proposition}
Under the setup above, for any candidate update vector \(s_t\),
\[
\Delta_t(s_t) := f(x_t) - f(x_t - \alpha_t P_t s_t)
\]
satisfies
\[
\Delta_t(s_t) = \alpha_t g_t^\top P_t s_t - \frac{\alpha_t^2}{2} s_t^\top P_t H P_t s_t,
\]
and therefore
\[
\Delta_t(s_t) \ge \alpha_t p_{\min} \|g_t\| \, \|s_t\| \cos(g_t, s_t) - \frac{\alpha_t^2}{2} \lambda_{\max}(H) p_{\max}^2 \|s_t\|^2.
\]
\end{proposition}

\begin{proof}
Since \(f\) is quadratic, its second-order Taylor expansion is exact:
\[
f(x_t - \alpha_t P_t s_t) = f(x_t) - \alpha_t \nabla f(x_t)^\top P_t s_t + \frac{\alpha_t^2}{2} s_t^\top P_t H P_t s_t.
\]

Using \(g_t = \nabla f(x_t)\), rearranging gives
\[
\Delta_t(s_t) = \alpha_t g_t^\top P_t s_t - \frac{\alpha_t^2}{2} s_t^\top P_t H P_t s_t.
\]

For the first term, since \(P_t \succeq p_{\min} I\),
\[
g_t^\top P_t s_t \ge p_{\min} g_t^\top s_t = p_{\min} \|g_t\| \, \|s_t\| \cos(g_t, s_t).
\]

For the second term, since \(H \preceq \lambda_{\max}(H) I\) and \(P_t \preceq p_{\max} I\),
\[
s_t^\top P_t H P_t s_t \le \lambda_{\max}(H) \, s_t^\top P_t^2 s_t \le \lambda_{\max}(H) p_{\max}^2 \|s_t\|^2.
\]

Substituting these two bounds into the exact identity yields
\[
\Delta_t(s_t) \ge \alpha_t p_{\min} \|g_t\| \, \|s_t\| \cos(g_t, s_t) - \frac{\alpha_t^2}{2} \lambda_{\max}(H) p_{\max}^2 \|s_t\|^2.
\]
This proves the claim.
\end{proof}

\subsection{Consequence for comparing two estimators}

To compare two candidate directions \(s_t^{(1)}\) and \(s_t^{(2)}\), define
\[
\overline{\Delta}_t(s) := \alpha_t p_{\min} \|g_t\| \, \|s\| \cos(g_t, s) - \frac{\alpha_t^2}{2} \lambda_{\max}(H) p_{\max}^2 \|s\|^2,
\]
the lower bound from the above proposition.

\begin{corollary}
Assume \(\|s_t^{(1)}\| = \|s_t^{(2)}\| = r\). If
\[
\cos(g_t, s_t^{(1)}) > \cos(g_t, s_t^{(2)}),
\]
then \(\overline{\Delta}_t(s_t^{(1)}) > \overline{\Delta}_t(s_t^{(2)})\).

More precisely,
\[
\overline{\Delta}_t(s_t^{(1)}) - \overline{\Delta}_t(s_t^{(2)}) = \alpha_t p_{\min} \|g_t\| \, r \, (\cos(g_t, s_t^{(1)}) - \cos(g_t, s_t^{(2)})).
\]
\end{corollary}

\begin{proof}
Since the two directions have the same norm \(r\), the quadratic penalty term in the lower bound is identical for both. Hence the difference between the two lower bounds is entirely determined by the cosine term:
\[
\overline{\Delta}_t(s_t^{(1)}) - \overline{\Delta}_t(s_t^{(2)}) = \alpha_t p_{\min} \|g_t\| \, r \, (\cos(g_t, s_t^{(1)}) - \cos(g_t, s_t^{(2)})).
\]
The result follows immediately.
\end{proof}

% This document was modified from the file originally made available by
% Pat Langley and Andrea Danyluk for ICML-2K. This version was created
% by Iain Murray in 2018, and modified by Alexandre Bouchard in
% 2019 and 2021 and by Csaba Szepesvari, Gang Niu and Sivan Sabato in 2022.
% Modified again in 2023 and 2024 by Sivan Sabato and Jonathan Scarlett.
% Previous contributors include Dan Roy, Lise Getoor and Tobias
% Scheffer, which was slightly modified from the 2010 version by
% Thorsten Joachims & Johannes Fuernkranz, slightly modified from the
% 2009 version by Kiri Wagstaff and Sam Roweis's 2008 version, which is
% slightly modified from Prasad Tadepalli's 2007 version which is a
% lightly changed version of the previous year's version by Andrew
% Moore, which was in turn edited from those of Kristian Kersting and
% Codrina Lauth. Alex Smola contributed to the algorithmic style files.

\clearpage
\section{Experimental Details and Hyperparameters}
\label{appendix:experimental_details}

In this section, we provide the full set of hyperparameters and experimental details used in our study.

\subsection{GLUE Benchmark Experiments}
For the classification tasks on the GLUE benchmark, we fine-tuned both RoBERTa-base and RoBERTa-large models. All classification experiments were executed on compute nodes equipped with a single NVIDIA L40S GPU. The common hyperparameters used across all GLUE datasets are as follows:
\begin{itemize}
    \item \textbf{Logical Batch Size}: 2000
    \item \textbf{Learning Rate}: \(5 \times 10^{-4}\)
    \item \textbf{Max Gradient Norm (clipping threshold \(C\))}: 10.0
    \item \textbf{Weight Decay}: 0.01
    \item \textbf{Total Training Steps}: 400
    \item \textbf{Privacy Target Epsilon (\(\epsilon\))}: 6.7
    \item \textbf{Privacy Target Delta (\(\delta\))}: \(10^{-5}\) (for all datasets except MNLI and QQP, which use \(10^{-6}\) due to their larger size)
    \item \textbf{Denoising Threshold Multiplier (\(\kappa\))}: 1.02
\end{itemize}

For the LoRA configuration, we used a rank of \(r=16\), \(\alpha=16\), and applied the adapters to the following target modules: \texttt{query}, \texttt{value}, \texttt{key}, \texttt{intermediate.dense}, and \texttt{output.dense}.

Each configuration in Table~\ref{tab:glue_results} was run five times. For two configurations, DP-LoRA on QQP with RoBERTa-base and our method on QQP with RoBERTa-large, one of the five runs was terminated by a hardware error, so these entries are computed from four runs. Confidence intervals throughout the paper are 95\% intervals of the mean, \(\bar{x} \pm t_{0.975,\,n-1}\, s/\sqrt{n}\), where \(s\) is the sample standard deviation over \(n\) runs.

\subsection{Generation Tasks}
For the generation tasks (E2E and DART), we fine-tuned the Llama-3.2-1B and Qwen-1.7B models. All generation experiments were executed on compute nodes equipped with four NVIDIA L40S GPUs. The hyperparameters used for these generation tasks are:
\begin{itemize}
    \item \textbf{Logical Batch Size}: 64
    \item \textbf{Learning Rate}: \(2 \times 10^{-4}\)
    \item \textbf{Max Gradient Norm (clipping threshold \(C\))}: 1.0
    \item \textbf{Weight Decay}: 0.01
    \item \textbf{Total Training Steps}: 400
    \item \textbf{Privacy Target Epsilon (\(\epsilon\))}: 5.4
    \item \textbf{Privacy Target Delta (\(\delta\))}: \(10^{-5}\)
\end{itemize}

For the LoRA configuration in these tasks, we used a rank of \(r=32\), \(\alpha=64\), and applied the adapters to the \texttt{q\_proj} and \texttt{v\_proj} modules. During the evaluation phase, text generation was performed using beam search with 10 beams, a maximum of 100 new tokens, a length penalty of 0.9, and a no-repeat n-gram size of 4.

\end{document}